%% file: main.tex
\documentclass{article}

    \PassOptionsToPackage{numbers,sort}{natbib}

\usepackage[preprint]{neurips_2019}

\usepackage[utf8]{inputenc} 
\usepackage[T1]{fontenc}    
\usepackage{hyperref}       
\usepackage{url}            
\usepackage{booktabs}       
\usepackage{amsfonts}       
\usepackage{nicefrac}       
\usepackage{microtype}      

\usepackage{amsmath}
\usepackage{amssymb}
\usepackage{subcaption}
\usepackage{adjustbox}
\usepackage{threeparttable}
\usepackage{multirow}
\usepackage{algorithm}
\usepackage{algorithmicx}
\usepackage{algpseudocode}
\usepackage{enumitem}

\usepackage{color}
\definecolor{darkred}{rgb}{0.8,0,0}
\definecolor{darkgreen}{rgb}{0,0.7,0}
\definecolor{darkblue}{rgb}{0,0.0,0.3}
\hypersetup{
  colorlinks,
  linkcolor=darkred,
  citecolor=darkgreen,
  urlcolor=darkblue,
  pdfinfo={
    Title={Federated Learning with Unbiased Gradient Aggregation and Controllable Meta Updating},
    Author={Xin Yao, Tianchi Huang, Rui-Xiao Zhang, Ruiyu Li, Lifeng Sun},
  }
}

\newcommand{\ours}{FedMeta w/ UGA}

\title{Federated Learning with Unbiased Gradient Aggregation and Controllable Meta Updating}

\author{
  Xin Yao\textsuperscript{\rm 1, 2}, Tianchi Huang\textsuperscript{\rm 1}, Rui-Xiao Zhang\textsuperscript{\rm 1}, Ruiyu Li\textsuperscript{\rm 2}, Lifeng Sun\textsuperscript{\rm *1} \\
  \textsuperscript{\rm 1}Department of Computer Science and Technology, Tsinghua University, Beijing, China \\
  \textsuperscript{\rm 2} Youtu X-Lab, Tencent, Shenzhen, China \\
  \texttt{\{yaox16,htc19,zhangrx17\}@mails.tsinghua.edu.cn} \\
  \texttt{royryli@tencent.com, sunlf@tsinghua.edu.cn}\\
}

\begin{document}

\maketitle

\begin{abstract}
Federated learning (FL) aims to train machine learning models in the decentralized system consisting of an enormous amount of smart edge devices.
Federated averaging (FedAvg), the fundamental algorithm in FL settings, proposes on-device training and model aggregation to avoid the potential heavy communication costs and privacy concerns brought by transmitting raw data.
However, through theoretical analysis we argue that 1) the multiple steps of local updating will result in gradient biases and 2) there is an inconsistency between the expected target distribution and the optimization objectives following the training paradigm in FedAvg.
To tackle these problems, we first propose an unbiased gradient aggregation algorithm with the keep-trace gradient descent and the gradient evaluation strategy.
Then we introduce an additional controllable meta updating procedure with a small set of data samples, indicating the expected target distribution, to provide a clear and consistent optimization objective.
Both the two improvements are model- and task-agnostic and can be applied individually or together.
Experimental results demonstrate that the proposed methods are faster in convergence and achieve higher accuracy with different network architectures in various FL settings.
\end{abstract}

\input{1-intro.tex}
\input{2-motivation.tex}
\input{3-method.tex}
\input{4-exp.tex}
\input{5-others.tex}

\section*{Acknowledgement}
This work is done when the first author visits Youtu X-Lab, Tencent, as a research intern.
This work is supported by the National Key R\&D Program of China (2018YFB1003703), the National Natural Science Foundation of China (61936011), as well as the Beijing Key Lab of Networked Multimedia (Z161100005016051).

\clearpage
\bibliographystyle{abbrv}
\bibliography{ref.bib}

\end{document}

%% file: 1-intro.tex
\section{Introduction}

With the advances in mobile technology, personal smart devices have become an indispensable part of modern life.
Meanwhile, massive IoT (Internet of Things) devices are expected to deploy in the next few years.
These devices are generating a tremendous amount of valuable data, from text messages to traffic status, which has the potential to bring intelligence to our daily life.
However, in traditional approaches, making full use of these data requires gathering data to the data centers and training machine learning models there, which is unrealistic in practice due to privacy concerns and the unreliable network transmission.

To get rid of the dilemma, federated learning (FL) \cite{konevcny2015federated,konevcny2016fo,mcmahan2017communication} proposes leveraging the massive decentralized computing resources to perform on-device training with their local data.
Federated averaging (FedAvg)~\cite{mcmahan2017communication}, the fundamental framework in FL settings, selects a part of clients for participating in training in each round, and then performs several epochs of local updating on the selected clients, and finally aggregates the local models or updates on the server to get the global model.
With this training paradigm, FedAvg protects the privacy of personal data and avoids the heavy communication costs.

FL raises several types of issues, including the system challenges (e.g., a massive number of edge clients with limited network connections), the statistic challenges (e.g., unbalanced and non-IID data distributions), and the data privacy preservation, which have attracted a lot of recent research interests.
To bring down the communication costs, \cite{konevcny2016fl,caldas2018expanding,sattler2019robust} propose structured model simplification and gradient compression methods to reduce the amount of data transmitted in each round; \cite{yao2018two,yao2019federated} propose adding additional modules or mechanisms to the on-device training, to accelerate the convergence and thus reduce the total communication rounds.
For the statistic challenges, \cite{li2019convergence} gives the convergence analysis of FedAvg;
\cite{sahu2018convergence} explores the federated optimization in heterogeneous systems;
\cite{ji2019learning} introduces attention mechanisms to the model aggregation to improve the performance for keyboard suggestion.
As for data privacy preservation, a lot of research efforts have been made in data encryption and security, including differential privacy guarantees~\cite{agarwal2018cpsgd}, secure frameworks~\cite{liu2018secure,cheng2019secureboost}, and adversarial attack~\cite{bhagoji2019analyzing} and defense~\cite{han2019robust}, etc.
However, only a few work~\cite{zhao2018federated,jeong2018communication} have made \emph{model- and task-agnostic} improvements over FedAvg in the general FL settings.


\begin{figure}[t]
    \centering
    \includegraphics[width=0.95\linewidth]{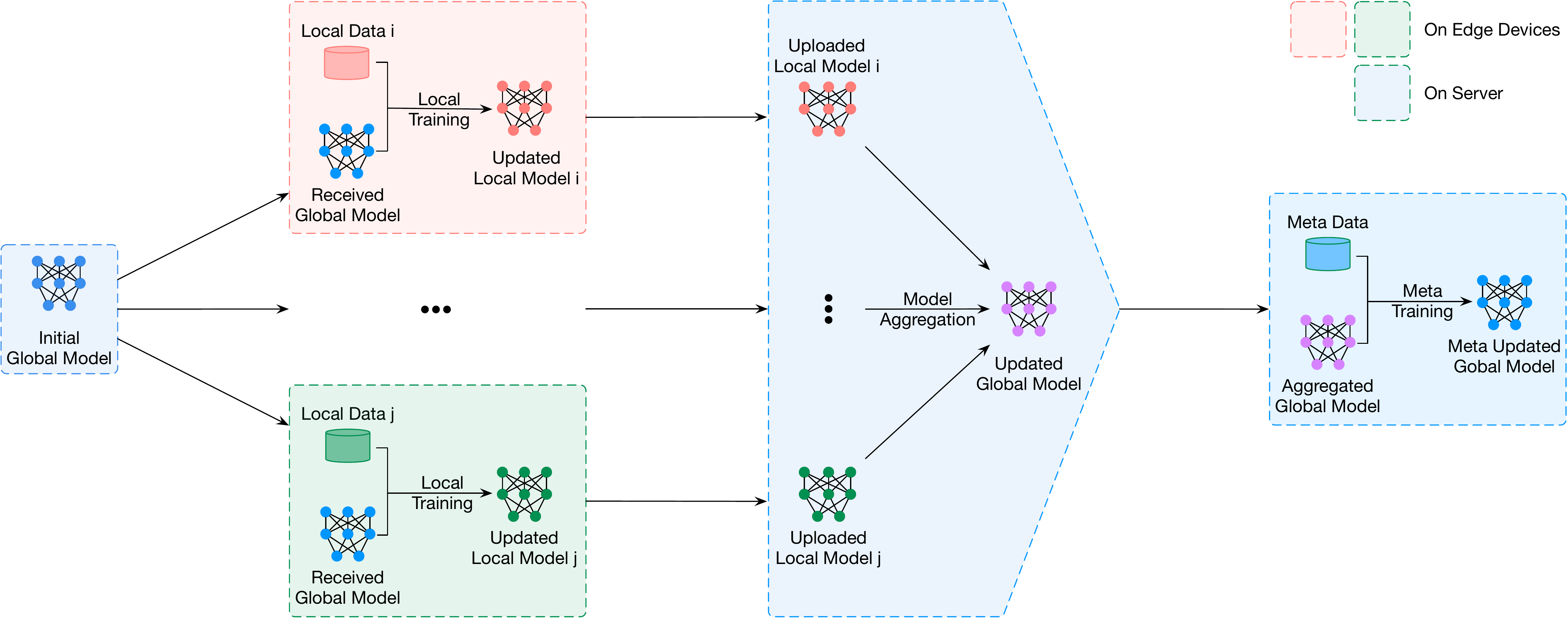}
    \caption{A typical round in FedMeta: model distributing, local training with either FedAvg~\cite{mcmahan2017communication} or UGA (Section \ref{sec:method_1}), model aggregating and meta updating. (Better viewed in color)}
    \label{fig:fedmeta}
\end{figure}

In this paper, we take a closer look at the FedAvg algorithm through theoretical analysis and argue that
1) the multiple steps of updating on clients will bring gradient biases to model aggregation (Section~\ref{sec:gradient_bias}) and 2) selecting a part of clients for participating in training in each round will result in an inconsistency between the optimization objectives and the real targeted distribution (Section~\ref{sec:inconsistency}).

To tackle these two problems, we first develop an unbiased gradient aggregation algorithm (UGA, Section~\ref{sec:method_1}) with the keep-trace gradient descent and gradient evaluation strategy.
Then we further introduce an additional meta updating procedure (FedMeta, Section~\ref{sec:method_2}) with a controllable meta training set on the server after model aggregation in each round.
Both the two improvements are model- and task-agnostic and can be applied individually or together.

We conduct experiments with various network architectures, including the convolutional neural networks~(CNNs) and the gated recurrent unit~(GRU) network, in both the IID and non-IID FL settings.
Results show the proposed methods are faster in convergence and achieve higher accuracy than the baselines, especially in non-IID FL settings.

To summarize, our contributions are as follows:
\begin{itemize}
    \item We develop an unbiased gradient aggregation algorithm for FL with the keep-trace gradient descent and gradient evaluation strategy, which is compatible with the existing FedAvg framework.
    \item We introduce an additional meta updating procedure after model aggregation on the server. It establishes a clear and consistent objective and guides the optimization of federated models in a controllable manner.
    \item Experiments with various network architectures in both IID and non-IID FL settings show the proposed methods are faster in convergence and achieve higher accuracy than the baselines.
\end{itemize}

%% file: 2-motivation.tex
\section{Problem Analysis and Motivation}

In this section, we will elaborate on the deficiencies in the vanilla FedAvg algorithm, which motivate this work.

The basic idea of FedAvg is derived from the distributed learning system consisting of parameter servers and computational workers.
Concretely, let us consider the FL system containing one single parameter server and $K$ computational workers.
At the beginning of round $t$, the parameter server distributes the model parameters $\omega_t$ to the workers.
Then each worker $k \in K$ computes one-step gradient $g_t^{k(1)} = \nabla_{\omega_t}\mathcal{L}_k(\omega_t;\mathcal{D}_k) = \nabla_{\omega_t}\sum_{(x_i, y_i) \in \mathcal{D}_k}\ell(\omega_t;x_i, y_i)$ where
$\mathcal{D}_k$ is the data distribution on client $k$ with $n_k = |\mathcal{D}_k|$ and $\ell$ is the loss function.
Next, the parameter server gathers all the gradients and applies the update with weighted average:
\begin{equation}
    \label{eq:1}
    \omega_{t+1} \leftarrow \omega_t - \eta\sum_{k \in K}\frac{n_k}{n}g_t^{k(1)}
\end{equation}
where $n = \sum_{k \in K}n_k$ and $\eta$ is the learning rate.
However, following the above updating paradigm, each of the $K$ workers has to communicate with the parameter server twice (receiving model parameters and sending gradients) in each round, which is a heavy burden in FL settings.

For this problem, FedAvg proposes two major improvements.
Firstly, they think Equation (\ref{eq:1}) is an equivalent to the weighted average of locally updated parameters $\omega_{t+1}^k$:
\begin{align}
\omega_{t+1} &\leftarrow \omega_t - \eta\sum_{k \in K}\frac{n_k}{n}g_t^{k(1)} \label{eq:fedavg1}\\
             & = \sum_{k \in K}\frac{n_k}{n}(\omega_t - \eta g_t^{k(1)}) \\
             & = \sum_{k \in K}\frac{n_k}{n}\omega_{t+1}^k
\end{align}
Thus they add more computation to each client by iterating the local update $\omega_{t}^{k(i)} = \omega_t^{k(i-1)} - \eta g_t^{k(i)}$ ($i$-th step, $\omega_t^{k(0)} = \omega_t$) multiple times before sending the parameters to server.
In other words, FedAvg reduces the overall communication rounds by increasing the local computations.

\subsection{Gradient Bias}
\label{sec:gradient_bias}

Nevertheless, Equation (\ref{eq:fedavg1}) makes sense because every $g_t^{k(1)}$ is the derivative of $\omega_t$ and thus the weighted average of $g_t^{k(1)}$ equals $g_t^{(1)}$, i.e., the gradients computed in a central way:
\begin{align}
g_t^{(1)} &= \nabla_{\omega_t}\mathcal{L}(\omega_t;\mathcal{D}) \\
  &= \sum_{k \in K}\frac{n_k}{n}\nabla_{\omega_t}\mathcal{L}_k(\omega_t;\mathcal{D}_k) \\
  &= \sum_{k \in K}\frac{n_k}{n}g_t^{k(1)} \label{eq:fedavg2}
\end{align}
where $\mathcal{D} = \sum_{k \in K}\frac{n_k}{n}\mathcal{D}_k$ is the overall data distribution.

We argue that Equation (\ref{eq:fedavg2}) \emph{does not hold with multiple steps of gradient descent on clients}.
For the sake of an intuitive explanation, we take a look at the second step of gradient descent $g_t^{(2)}$ and $g_t^{k(2)}$:
\begin{align}
g_t^{(2)}   &= \nabla_{\omega_t^{(1)}}\mathcal{L}(\omega_t^{(1)};\mathcal{D}) \\
g_t^{k(2)} &= \nabla_{\omega_t^{k(1)}}\mathcal{L}(\omega_t^{k(1)};\mathcal{D}_k)
\end{align}
where $\omega_t^{(1)}$ and $\omega_t^{k(1)}$ denote the centrally and locally updated version of $\omega_t$ respectively after one-step gradient descent.
Obviously, Equation (\ref{eq:fedavg2}) does not hold here because $g_t^{(2)}$ and $g_t^{k(2)}$ are the derivatives of different parameters, i.e., $\omega_t^{(1)}$ and $\omega_t^{k(1)}$, respectively.
We denote the gap between $g_t$ and $\sum_{k \in K}\frac{n_k}{n}g_t^k$ as the \emph{gradient bias}, which is rather small at the beginning but accumulates as the local updating step increases and finally harms the performance of federated models, especially in non-IID FL settings~\cite{zhao2018federated}.

\subsection{Inconsistent Optimization Objectives}
\label{sec:inconsistency}

The second major improvement proposed by FedAvg is selecting a part of workers (or clients in FL settings) for performing computation in each round, i.e., replacing $K$ with $C \cdot K$, where $C$ is the fraction of clients.
They have shown that a small $C$ (e.g., $C=0.1$) can lead to convergence.
However, it will bring another problem, i.e., \emph{the inconsistency between the optimization objectives and the real target distribution}~\cite{mohri2019agnostic}.

On round $t$, FedAvg actually trains the global model to minimize the empirical loss on the distribution $\mathcal{D}_{S_t}$:
\begin{align}
\mathcal{D}_{S_t} = \sum_{k \in S_t}\frac{n_k}{n_{S_t}}\mathcal{D}_k,\qquad n_{S_t} = |\mathcal{D}_{S_t}|
\end{align}
$S_t$ is the random set of $C \cdot K$ clients in round $t$.
The aggregated gradient $g_t$ in Equation~(\ref{eq:fedavg2}) indicates the descent direction on $\mathcal{D}_{S_t}$ instead of the expected $\mathcal{D}$.

There are two main reasons for the inconsistency.
On the one hand, $\mathcal{D}_{S_t}$ varies between rounds, which results in the lack of a clear and consistent optimization objective.

On the other hand, there is a gap between $\mathcal{D}_{S_t}$ and the real target distribution $\mathcal{D}$, which may be caused by the biased selection of clients.
For example, when developing a model for mobile phone users, FedAvg performs model training directly on mobile devices and thus requires the participants to have certain computation power, which is seldom met on cheap devices.
As a result, the federated model would be trained with biased data distributions mainly come from the expensive mobile phones.
Such a model is not qualified to deploy in the broad and general domain formed by both expensive and cheap (usually more in quantity) devices.

%% file: 3-method.tex
\section{Method}

To tackle the aforementioned problems, we propose the unbiased gradient aggregation~(UGA) and controllable meta updating~(FedMeta), both of which are not limited to specific models or tasks, and can be applied individually or together.

\subsection{Unbiased Gradient Aggregation}
\label{sec:method_1}

The gradient bias can be briefly summarized as that the gradients $g_t$ and $g_t^k$ are calculated against different model parameters, and thus the weighted average $g_t = \sum_{k \in K}\frac{n_k}{n}g_t^k$ in FedAvg does not hold with multiple steps of gradient descent.
A straightforward solution is to calculate the gradients against $\omega_t$ instead of $\omega_t^{k(i)}$ in the $i$-th step of round $t$ on client $k \in S_t$.
However, such a solution violates the gradient descent optimization rule on clients, and thus its convergence cannot be guaranteed.
What's worse, it requires calculating the high-order derivatives in every step and thus brings heavy computation and storage (of intermediate variables) costs, which are unbearable for smart edge devices.

To get out of this dilemma, we develop a novel gradient evaluation strategy that is able to calculate the gradients in an unbiased and computation-efficient manner.
Briefly taking an on-device training procedure with $E$ local epochs for example, we perform the keep-trace gradient descent optimization for the first $E - 1$ epochs, and then evaluate gradients using the whole local data in the last epoch.

\begin{algorithm}[t]
    \caption{Unbiased Gradient Aggregation}
    \label{algorithm1}
    \begin{algorithmic}[1]
        \Statex \textbf{Server Executes:}
        \State Initialize $\omega_0$
        \For {each round $t = 0, 1, ...$}
        \State $m \leftarrow$ max$(C \cdot K, 1)$
        \State $S_t \leftarrow$ (random set of $m$ clients)
            \For {each client $k \in S_t$} \textbf{in parallel}
                \State $g_t^k \leftarrow$ \textbf{ClientUpdate}($k, \omega_t$)
            \EndFor
            \State $\omega_{t+1} \leftarrow \omega_t - \eta_g \sum_{k \in S_t} \frac{n_k}{n_{S_t}} g_t^k$
            \Comment{\textit{Equation (\ref{eq:gradient_aggregation})}}
        \EndFor
    \end{algorithmic}
    \begin{algorithmic}[1]
        \Statex \textbf{ClientUpdate($k, \omega_t$):}
        \Comment{\textit{Run on client $k$}}
        \For {$i$ in the total steps of the first $E-1$ epochs}
            \State $\omega_t^{k(i)} \leftarrow \omega_t^{k(i-1)} - \eta g_t^{k(i)}$
            \Comment{with \textit{Keep-trace GD}}
        \EndFor
        \State $g_t^k = \nabla_{\omega_t} \mathcal{L}(\omega_t^k; \mathcal{D}_k)$
        \Comment{\textit{Equation (\ref{eq:gradient_evaluation})}}
        \State return $g_t^k$ to server
    \end{algorithmic}
\end{algorithm}

\subsubsection{Keep-trace Gradient Descent}


In the $i$-th step of updating on client $k$ in round $t$, vanilla gradient descent will execute
\begin{equation}
\label{eq:vanilla_gd}
\omega_t^{k(i)} = \omega_t^{k(i-1)} - \eta g_t^{k(i)}
\end{equation}
and keep just $\omega_t^{k(i)}$ as the initial state for the next step of updating.
Denoting $\mathcal{B}_t^{k(i)}$ as the batch of examples in the $i$-th step on client $k$ in round $t$, we have:
\begin{equation}
\label{eq:keep_trace}
g_t^{k(i)} = \nabla_{\omega_t^{k(i-1)}} \mathcal{L}(\omega_t^{k(i-1)}; \mathcal{B}_t^{k(i)})
\end{equation}
Notice that $g_t^{k(i)}$ is a function of $\omega_t^{k(i-1)}$ and thus $\omega_t^{k(i)}$ is also a function of $\omega_t^{k(i-1)}$, i.e., $\omega_t^{k(i)} = f_{k(i)}(\omega_t^{k(i-1)})$.
Instead of treating $g_t^{k(i)}$ as numerical values and Equation (\ref{eq:vanilla_gd}) as a numerical computation, we \emph{keep the functional relation} between $\omega_t^{k(i)}$ and $\omega_t^{k(i-1)}$, the model parameters of adjacent steps, when updating parameters.
For the first $E-1$ epochs, we conduct the forward and backward calculations through the model multiple times, and record the whole computational history, which is termed as the \emph{keep-trace gradient descent}.
The multiple epochs of local updating, in line with the vanilla FedAvg, is used to reduce the communication costs by increasing local computation.

\subsubsection{Gradient Evaluation}

After the $E-1$ epochs of local updating, we will finally get $\omega_t^k = h_k(\omega_t)$ according to the recursive relations, i.e., the functional relation between the final updated model parameters and the initial ones on client $k$.
Then in the last epoch, we evaluate $\omega_t^k$ on the whole client data and calculate the gradient against $\omega_t$, the \emph{shared} initial model parameters, by unrolling the computational history recorded in the above keep-trace gradient descent:
\begin{equation}
\label{eq:gradient_evaluation}
g_t^k = \nabla_{\omega_t} \mathcal{L}(\omega_t^k; \mathcal{D}_k)
\end{equation}
Since all the $g_t^k$ for $k \in S_t$ are the derivatives of $\omega_t$, we can aggregate the gradients on the parameter server in an unbiased way using:
\begin{equation}
\label{eq:gradient_aggregation}
\omega_{t+1} \leftarrow \omega_t - \eta_g \sum_{k \in S_t} \frac{n_k}{n_{S_t}} g_t^k
\end{equation}
where $\eta_g$ is the step size for gradient aggregation.
With the keep-trace gradient descent and the subsequent gradient evaluation strategy, the computation of high-order derivatives is only required on the last epoch, which is much more computation- and storage-efficient.

Here, we have a simple two-step example for explaining how the keep-trace gradient descent and the gradient evaluation strategy work.
First, we start from $\omega_t^{k(0)}$ and perform one step SGD to get $\omega_t^{k(1)}$, a function of $\omega_t^{k(0)}$:
\begin{align}
\omega_t^{k(1)} &= \omega_t^{k(0)} - \eta g_t^{k(1)} \\
                &= \omega_t^{k(0)} - \eta \nabla_{\omega_t^{k(0)}} \mathcal{L}(\omega_t^{k(0)}; \mathcal{B}_t^{k(1)}) \label{eq:w1}
\end{align}
Then we perform one more step keep-trace gradient descent with data batch $\mathcal{B}_t^{k(2)}$ to get $\omega_t^{k(2)}$:
\begin{align}
\omega_t^{k(2)} &= \omega_t^{k(1)} - \eta g_t^{k(2)} \\
            &= \omega_t^{k(1)} - \eta \nabla_{\omega_t^{k(1)}} \mathcal{L}(\omega_t^{k(1)}; \mathcal{B}_t^{k(2)}) \label{eq:w2}
\end{align}
Here we can replace $\omega_t^{k(1)}$ in Eq. (\ref{eq:w2}) with Eq. (\ref{eq:w1}) and get that $\omega_t^{k(2)}$ is also a function of $\omega_t^{k(0)}$ according to the recursive relation.
Then we evaluate it with data batch $\mathcal{B}_t^{k(3)}$ to compute the gradient $g_t^k$ with respect to $\omega_t^{k(0)}$:
\begin{align}
g_t^k &= \nabla_{\omega_t^{k(0)}} \mathcal{L}(\omega_t^{k(2)}; \mathcal{B}_t^{k(3)})
\end{align}
Finally we get all the $g_t^k$ for $k \in S_t$ are the derivatives of $\omega_t$.

The improved FedAvg with unbiased gradient aggregation~(UGA) is summarized as Algorithm \ref{algorithm1}.

\begin{algorithm}[t]
    \caption{FedMeta}
    \label{algorithm2}
    \begin{algorithmic}[1]
        \Statex \textbf{Server Executes:}
        \State Initialize $\omega_0$
        \For {each round $t = 0, 1, ...$}
        \State $m \leftarrow$ max$(C \cdot K, 1)$
        \State $S_t \leftarrow$ (random set of $m$ clients)
            \For {each client $k \in S_t$} \textbf{in parallel}
                \State $g_t^k \leftarrow$ \textbf{ClientUpdate}($k, \omega_t$)
                \Statex \qquad \textit{// Compatible with both FedAvg and Algorithm \ref{algorithm1}}
            \EndFor
            \State $\omega_{t+1} \leftarrow \omega_t - \eta_g \sum_{k \in S_t} \frac{n_k}{n_{S_t}} g_t^k$
            \Comment{\textit{Equation (\ref{eq:gradient_aggregation})}}
            \State $\omega_{t+1}^{meta} = \omega_{t+1} - \eta_{meta} \nabla_{\omega_{t+1}} \mathcal{L}(\omega_{t+1};\mathcal{D}_{meta})$
            \Statex \quad \textit{// Equation (\ref{eq:meta_update})}
        \EndFor
    \end{algorithmic}
\end{algorithm}

\subsection{Controllable Meta Updating}
\label{sec:method_2}

The inconsistency between the target distribution and the optimization objectives is caused by two factors:
1) $\mathcal{D}_{S_t}$ varies between rounds because different parts of clients are selected for performing computation in each round;
2) there is a gap between the selected $\mathcal{D}_{S_t}$ and the real target distribution $\mathcal{D}$.
In summary, we lack a clear and consistent objective.

To tackle this problem, we introduce an additional meta updating procedure (FedMeta) with a small set of data samples $\mathcal{D}_{meta}$ on the parameter server after model aggregation in each round.
Note that FedMeta is compatible with both UGA and the client updating strategy in the vanilla FedAvg.


The whole optimization process of FedMeta in round $t$ can be described as:
\emph{we want to optimize $\omega_t$, after the client updating on $C \cdot K$ clients and the gradient aggregation on the server to obtain $\omega_{t+1}$, the network parameter that performs well on the meta training set $\mathcal{D}_{meta}$}.
A representative round in FedMeta is illustrated in Fig. \ref{fig:fedmeta}.

This is a \emph{two-stage optimization} that contains: 1) the inner loop optimization on $C \cdot K$ clients using Equation (\ref{eq:keep_trace}), (\ref{eq:gradient_evaluation}) and (\ref{eq:gradient_aggregation});
and 2) the outer loop optimization on the server, i.e., the meta updating procedure, using:
\begin{equation}
\label{eq:meta_update}
\omega_{t+1}^{meta} = \omega_{t+1} - \eta_{meta} \nabla_{\omega_{t+1}} \mathcal{L}(\omega_{t+1};\mathcal{D}_{meta})
\end{equation}
where $\eta_{meta}$ is the meta learning rate.
The pseudo-code of FedMeta is shown in Algorithm \ref{algorithm2}.

It is worth noting that in this two-stage optimization, the whole training process has a clear and consistent objective, i.e., the performance on the meta training set $\mathcal{D}_{meta}$, which solves the problem in Section \ref{sec:inconsistency} but inevitably depends too heavily on the selection of $\mathcal{D}_{meta}$.

\subsubsection{The Role of $\mathcal{D}_{meta}$ and Privacy Concerns}

In a common situation, $\mathcal{D}_{meta}$ could be an IID subset of the overall data distribution $D$.
In practice, it could be acquired by the data voluntarily shared by some users, or by recruiting some users for participating in the insider program or testing the beta versions of federated applications.
These methods for constructing $\mathcal{D}_{meta}$ do not violate the privacy protection principles and have been adopted in some previous studies \cite{jeong2018communication,zhao2018federated}.
In particular, how to construct an appropriate $D_{meta}$ is not the focus of this paper.

\subsubsection{Controllable Federated Models}

Further, from another point of view, \emph{$\mathcal{D}_{meta}$ offers a way to pertinently control the behavior of federated models}.
In the vanilla FedAvg, the server (or the developers behind) cannot control what kind of model is finally trained by the system, which brings a lot of trouble to the model tuning and therefore limits the wide application of the algorithm.
For example, since the convergence of training federated models requires a great many communication rounds, you have to wait for another long time if you are not satisfied with the previous model and decide to retrain.
Different from that in FedAvg, the federated model is always optimized towards a better performance on the meta training set $\mathcal{D}_{meta}$ in FedMeta.
In other words, what kind of $\mathcal{D}_{meta}$ you offer, the corresponding federated model is trained.

In fact, there is no necessary connection between the meta training set $\mathcal{D}_{meta}$ and the overall data distribution $\mathcal{D}$.
Instead, $\mathcal{D}_{meta}$ should be explicitly chosen according to the real targets, which is a powerful tool for reducing the biases and unfairness of federated models.
For example, the overall data distribution $\mathcal{D}$ may contain some prejudices of gender, race or wealth~\cite{buolamwini2018gender,chen2018my,zou2018ai,olteanu2019social}, but we could build a better meta training set $\mathcal{D}_{meta}$ to guide the optimization of federated models towards an unbiased and fair manner.
In such a situation, the inner loop optimization on clients is considered as pre-training while the outer loop optimization on the server, i.e., the meta updating, is considered as fine-tuning on the meta training set $\mathcal{D}_{meta}$ that indicates the expected targets.

%% file: 4-exp.tex
\section{Experiments}

We evaluate the proposed methods with various network architectures in both the IID and non-IID FL settings.
Concretely, for the IID FL setting, we follow the setup in \cite{zhao2018federated} and manually split CIFAR-10~\cite{krizhevsky2009learning} into several subsets;
for the non-IID settings, we adopt FEMNIST and Shakespeare from a recently proposed FL benchmark LEAF~\cite{caldas2018leaf}.

The combined method of UGA and FedMeta, denoted as FedMeta w/ UGA, is compared with the following algorithms:
\begin{itemize}
    \item FedAvg: the conventional FL algorithm and framework proposed in~\cite{mcmahan2017communication}.
    \item FedProx~\cite{li2018federated}: it uses a proximal term to improve the performance of FedAvg in heterogeneous networks.
    \item FedShare~\cite{zhao2018federated}: it shares a small public dataset among clients to alleviate the weight divergence in FL.
\end{itemize}


We follow some notations in FedAvg and other FL algorithms: $B$, the local training batch size; $E$, the number of local training epochs; and $C$, the fraction of clients selected in each round.

We implement the proposed methods and the compared ones with PyTorch~\cite{NEURIPS2019_9015}.
Specially, the keep-trace gradient descent is implemented by creating computation graphs for $g_t^{k(i)}$ in Equation~(\ref{eq:keep_trace}) during the automatic differentiation.

Since we propose two improvements, i.e., UGA and FedMeta, to tackle the deficiencies in FedAvg, we conduct ablation studies to take a closer look at them separately in Section \ref{sec:ab}.

\begin{figure*}[t]
    \centering
    \begin{subfigure}{0.325\linewidth}
        \centering
        \includegraphics[width=\linewidth]{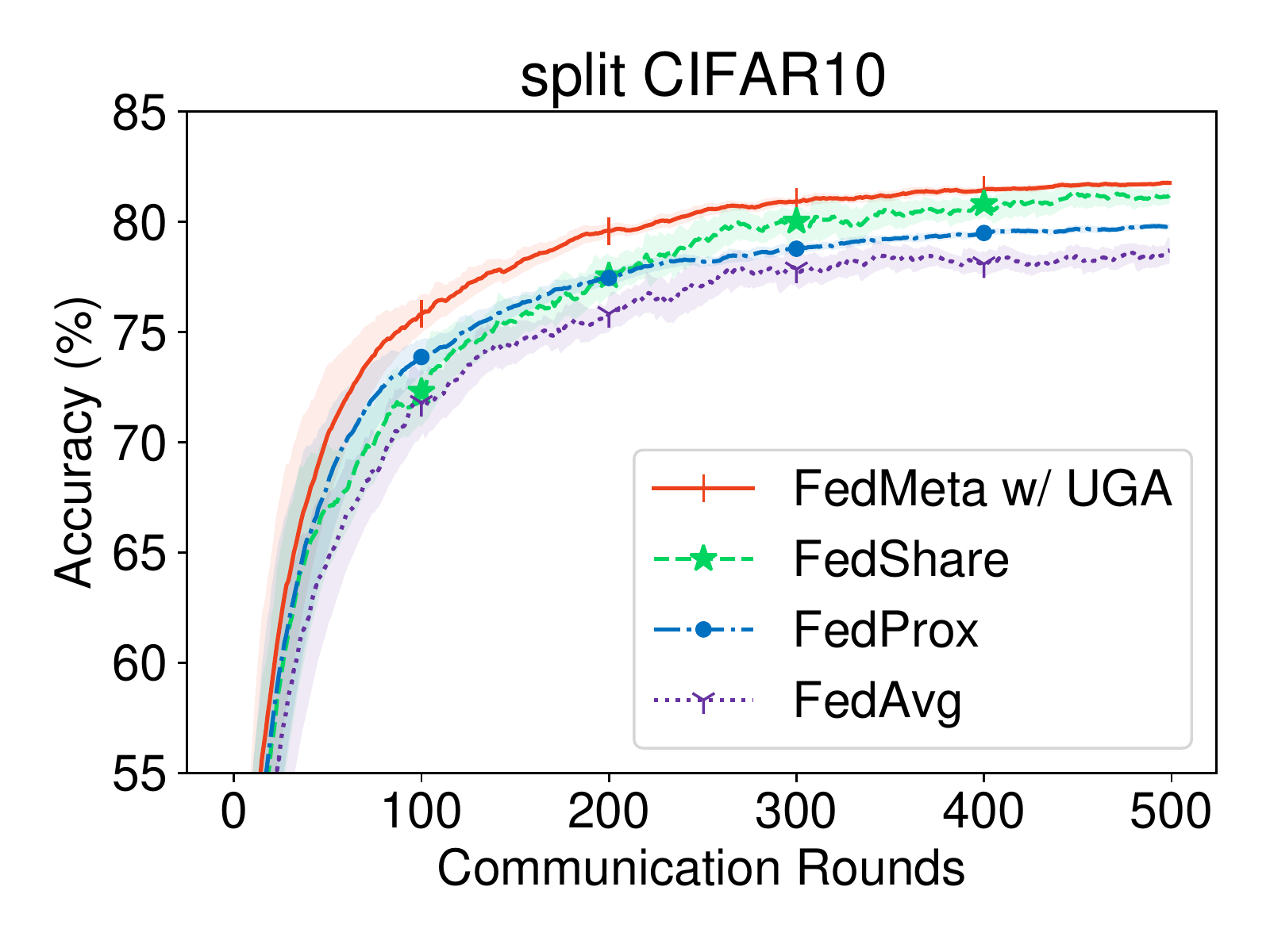}
        \includegraphics[width=\linewidth]{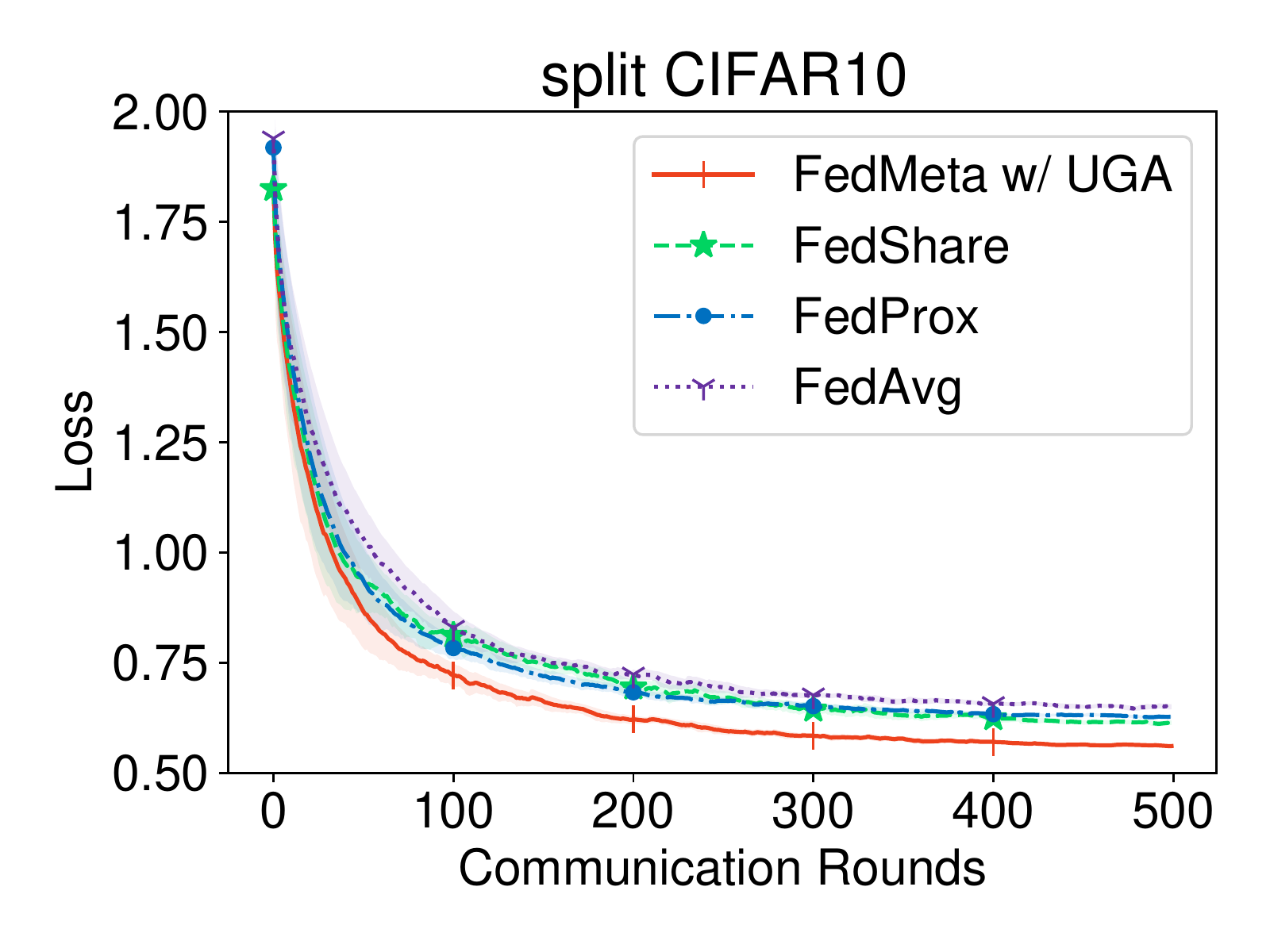}
        \caption{$E=2, B=64$}
        \label{fig:cifar_a}
    \end{subfigure}
    \begin{subfigure}{0.325\linewidth}
        \centering
        \includegraphics[width=\linewidth]{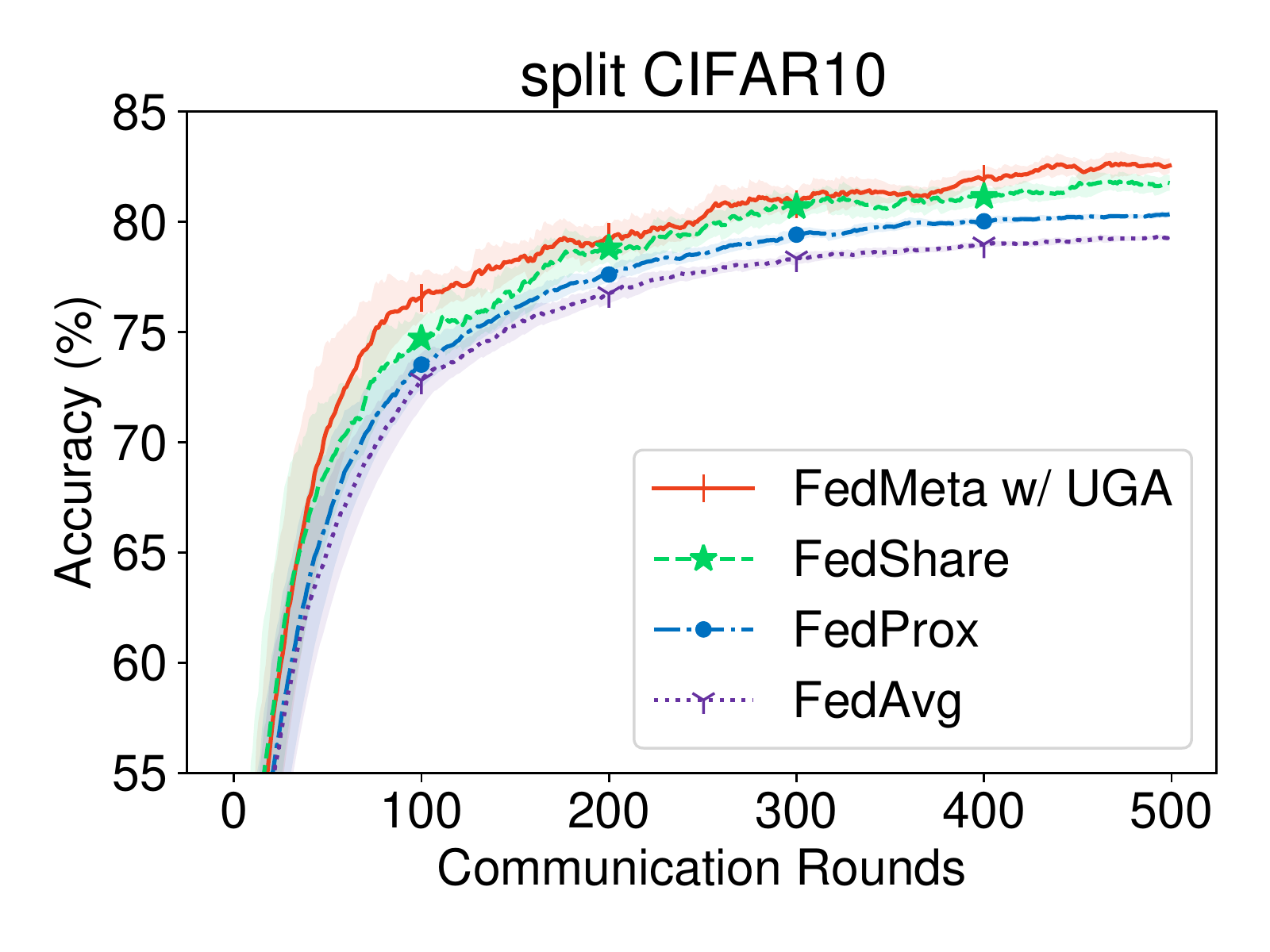}
        \includegraphics[width=\linewidth]{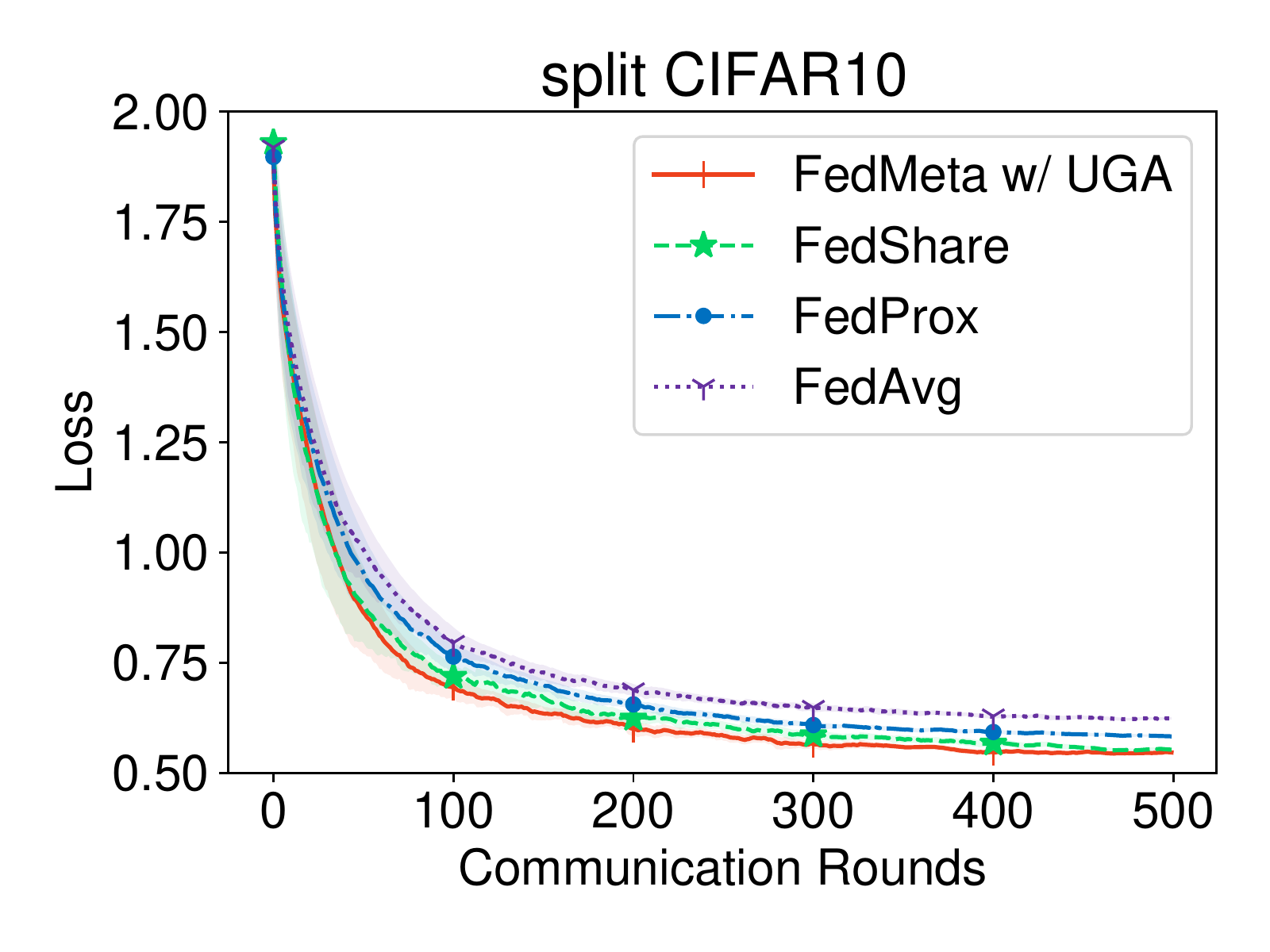}
        \caption{$E=2, B=128$}
        \label{fig:cifar_b}
    \end{subfigure}
    \begin{subfigure}{0.325\linewidth}
        \centering
        \includegraphics[width=\linewidth]{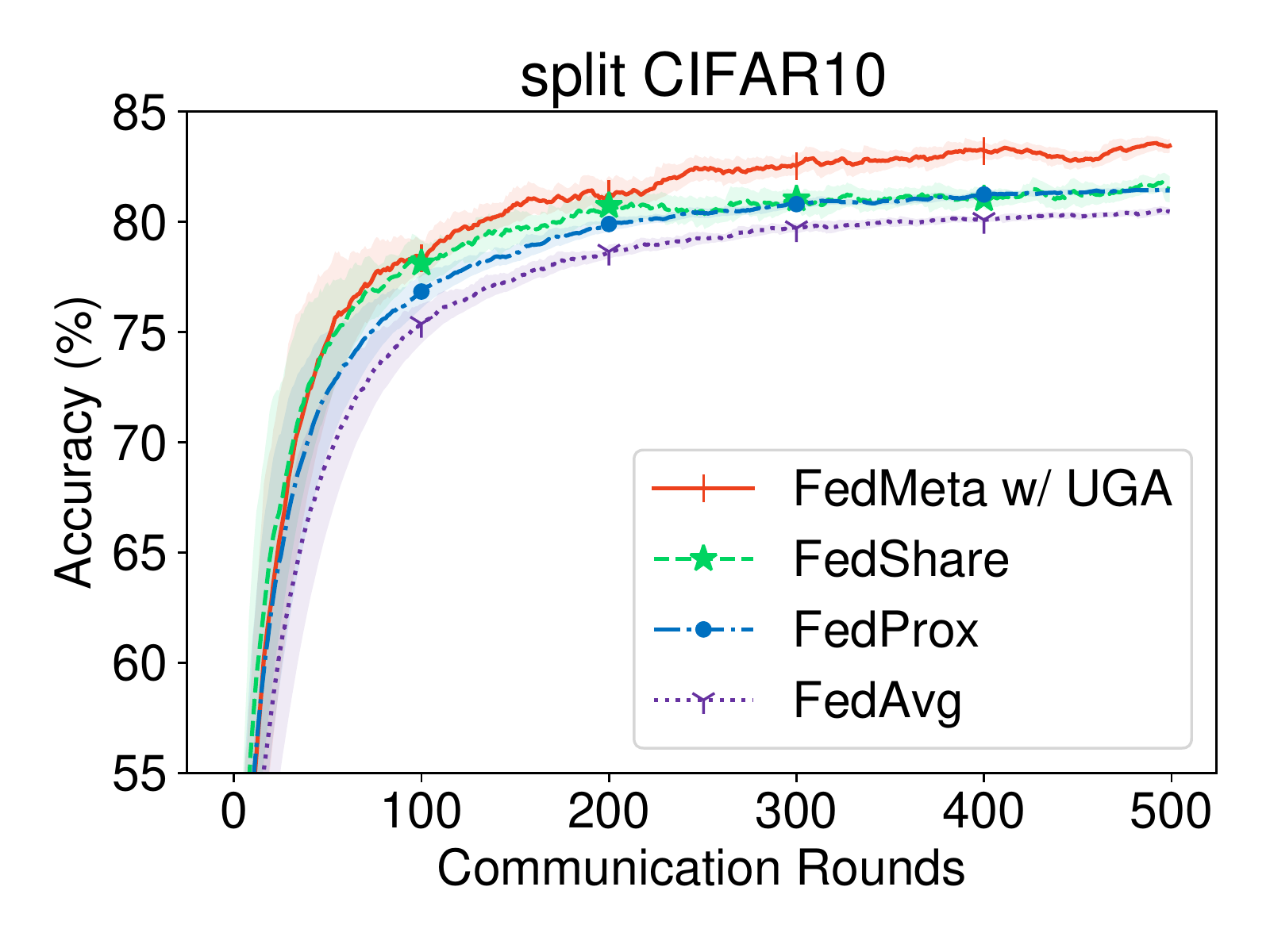}
        \includegraphics[width=\linewidth]{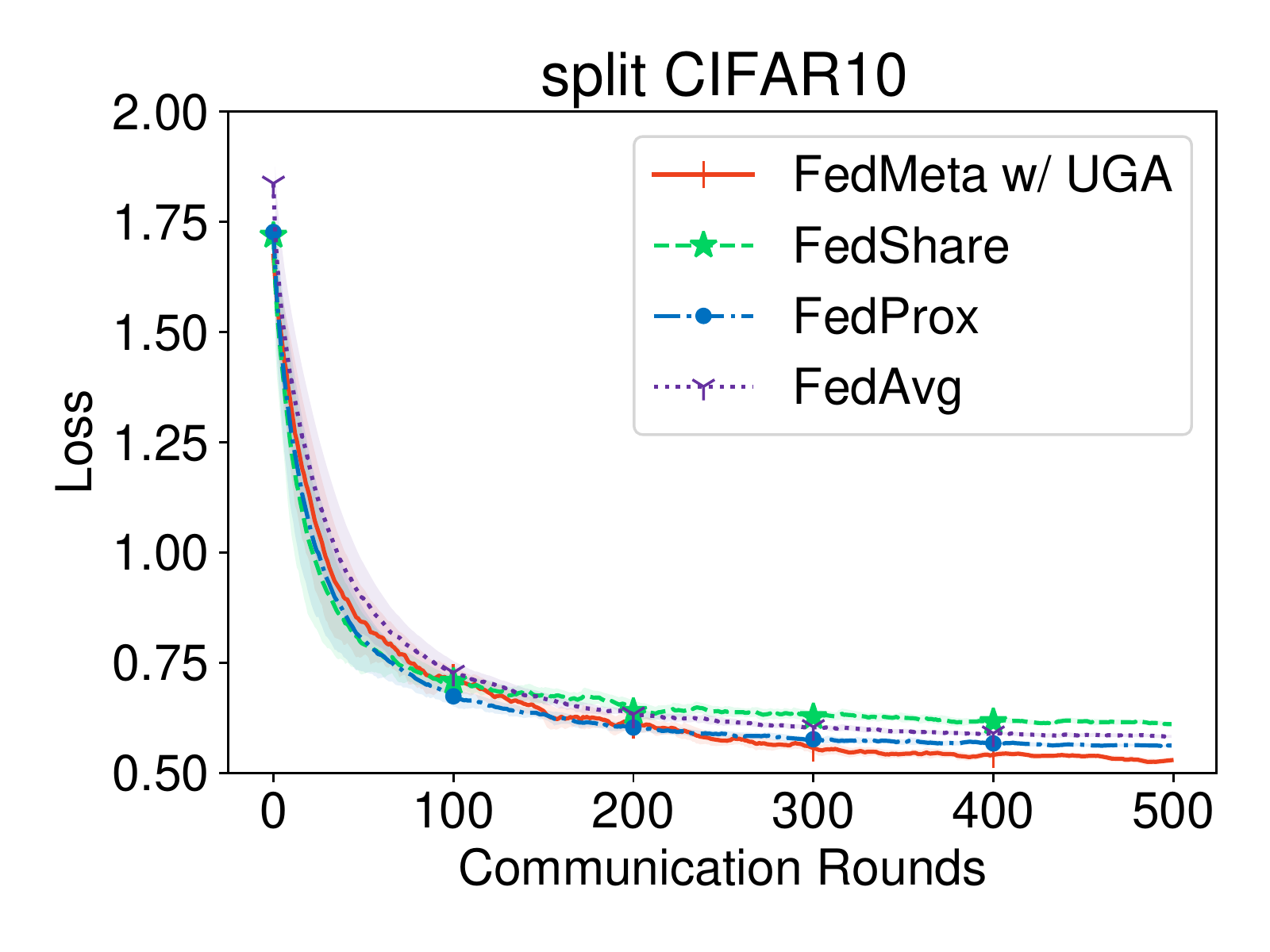}
        \caption{$E=5, B=128$}
        \label{fig:cifar_c}
    \end{subfigure}
    \caption{Test accuracy (upper row) and loss (lower row) over communication rounds of FedMeta w/ UGA compared to FedShare, FedProx and FedAvg with the CNN model on split CIFAR-10 (IID): (a) $E=2, B=64$; (b) $E=2, B=128$; (c) $E=5, B=128$. (Better viewed in color)}
    \label{fig:cifar}
\end{figure*}

\subsection{CNN Model on Split CIFAR-10 (IID)}
\label{sec:cifar}

\subsubsection{Dataset}
The 50,000 training images in CIFAR-10 are partitioned into 10 clients with each client randomly assigned a uniform distribution over the 10 classes.
Additionally, we randomly sample 1\% of the total images as the sharing data for FedShare and the meta training set for FedMeta.

\subsubsection{Model}
We use the same CNN architecture as FedAvg:
two 5$\times$5 convolution layers (both with 64 channels, each followed by a ReLU activation and 3$\times$3 max pooling with stride size 2), two fully connected layers (with 384 and 192 units respectively, each followed by a ReLU activation and random dropout) and a final softmax output layer.

\subsubsection{Hyper parameters}
$C=0.2$ (two clients) is fixed for experiments on split CIFAR-10.
We use the stochastic gradient descent (SGD) optimizer with the learning rate $\eta=0.002 (=\eta_{meta})$ and a decay rate $=0.992$ per communication round.
The coefficient for the proximal term in FedProx is set to $2\times10^{-4}$.

\begin{table}[]
\caption{The convergence accuracy of all the compared methods in split CIFAR-10 (IID).}
\label{table:cifar}
\centering
\begin{threeparttable}
\begin{tabular}{@{}lccc@{}}
\toprule
split CIFAR-10 & $E$=2, $B$=64  & $E$=2, $B$=128 & $E$=5, $B$=128   \\ \midrule
FedAvg         & 78.57          & 79.27          & 80.45          \\
FedProx        & 79.77          & 80.31          & 81.44          \\
FedShare       & 81.19          & 81.75          & 81.66          \\ \midrule
FedMeta w/ UGA & \textbf{81.74} & \textbf{82.58} & \textbf{83.43} \\ \bottomrule
\end{tabular}
\begin{tablenotes}
\item[*] Bold fonts indicate better performances.
\end{tablenotes}
\end{threeparttable}
\end{table}

\subsubsection{Results}
The convergence curves of the FedMeta w/ UGA over FedShare, FedProx, and FedAvg with different $E$ or $B$ are illustrated in Fig.~\ref{fig:cifar}.
In all the three experiments, FedProx only slightly outperforms FedAvg, which shows that the proximal term has limited effect in the IID FL settings.
FedShare improve the performance of FedAvg with the help of the globally shared dataset when $E$ is small, as shown in Fig. \ref{fig:cifar_a} and \ref{fig:cifar_b}.
As a contrast, FedMeta w/ UGA is faster in convergence and achieves better convergence performance than FedProx and FedAvg with a large margin under all circumstances.
It also outperforms FedShare without data sharing among clients.
We further summary the accuracy values in Table \ref{table:cifar} to see the results more clearly.
\ours outperforms FedAvg with the improvement on accuracy by more than 3 percentage points on average.

It is worth noting that when there are more updating steps (smaller $B$ or larger $E$) on the local clients, the improvements made by the proposed method are more significant.
For example, when $B$ is smaller (Fig. \ref{fig:cifar_a} vs. \ref{fig:cifar_b}), FedMeta w/ UGA shows a much faster convergence speed, with more than 50\% reduction in the needed communication rounds to reach the accuracy of 75\% compared to FedAvg, from 79 rounds to 39 rounds; and when $E$ is larger (Fig. \ref{fig:cifar_c}), FedMeta w/ UGA outperforms FedProx and FedProx in accuracy by about 2 percentage points, which is a more significant improvement than those in Fig. \ref{fig:cifar_a} and \ref{fig:cifar_b}.
This can be due to the accumulated gradient biases as the updating step on clients increases, which exacerbates the weight divergence~\cite{zhao2018federated} between the local and global models.
What's worse, though changes in the optimization objectives between rounds are small, they will bring instability to the optimization process and finally result in poor performances.

\begin{figure*}[t]
    \centering
    \begin{subfigure}{0.32\linewidth}
        \centering
        \includegraphics[width=\linewidth]{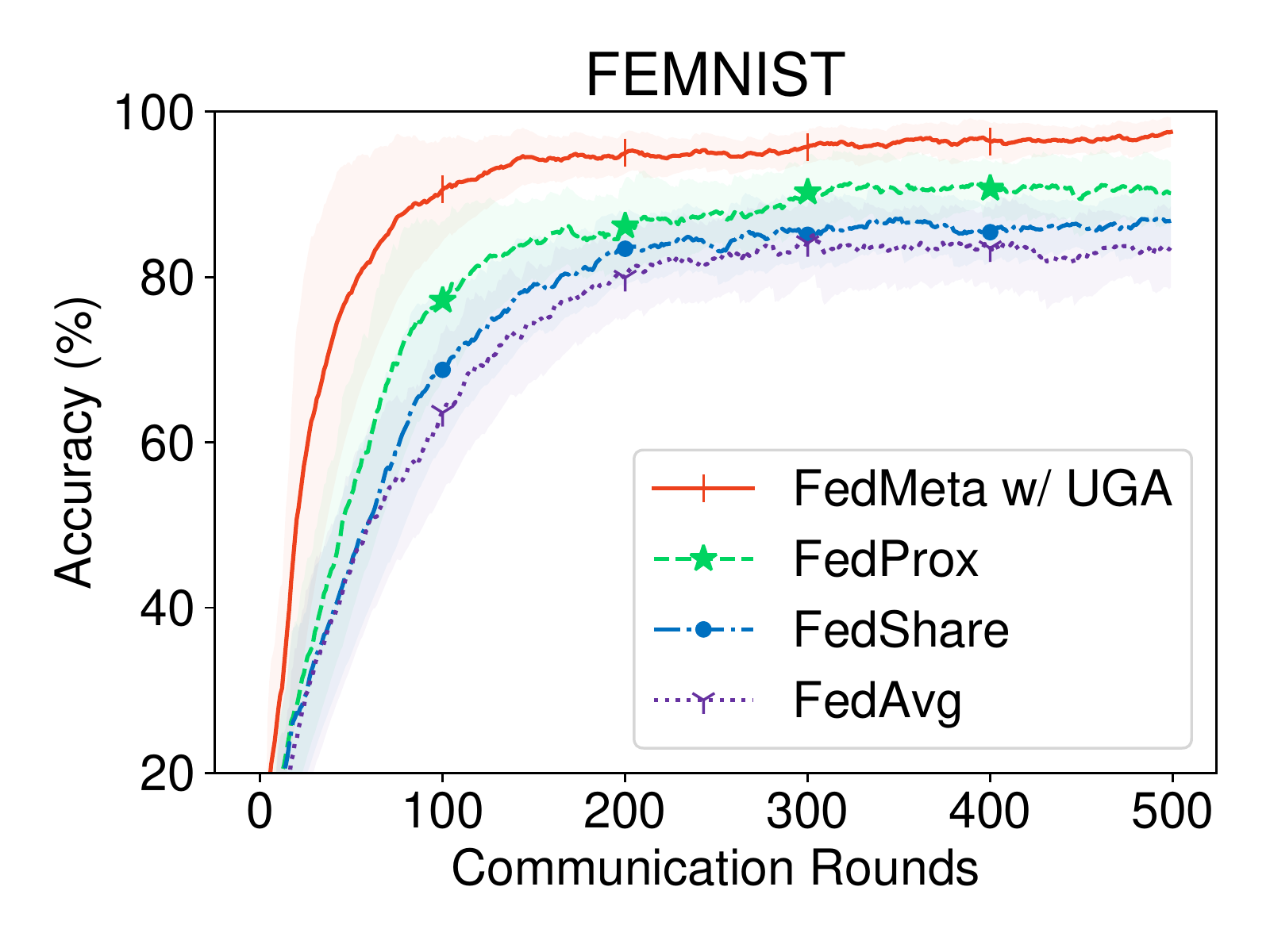}
        \includegraphics[width=\linewidth]{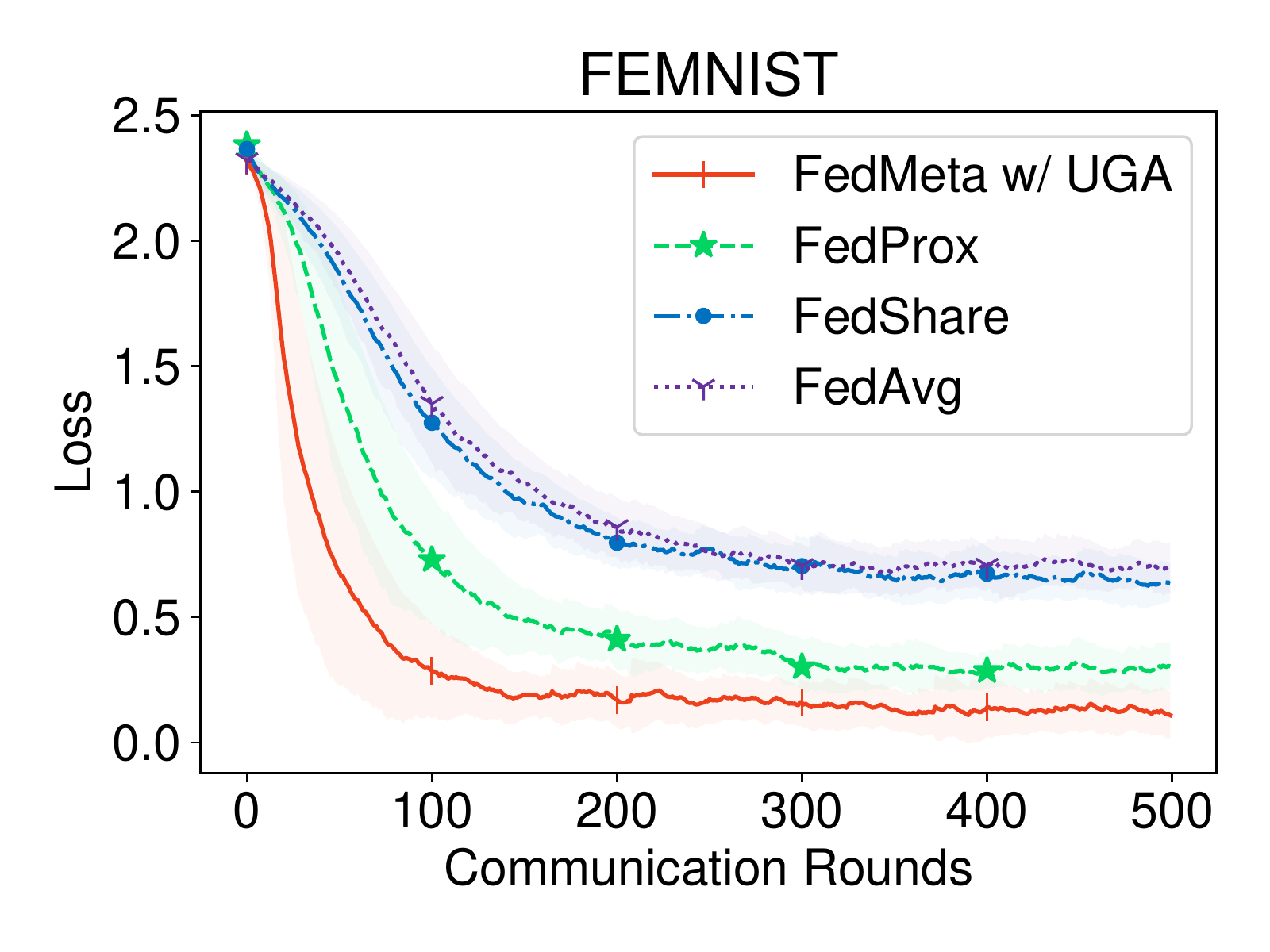}
        \caption{$E=2, B=64$}
        \label{fig:femnist_a}
    \end{subfigure}
    \begin{subfigure}{0.32\linewidth}
        \centering
        \includegraphics[width=\linewidth]{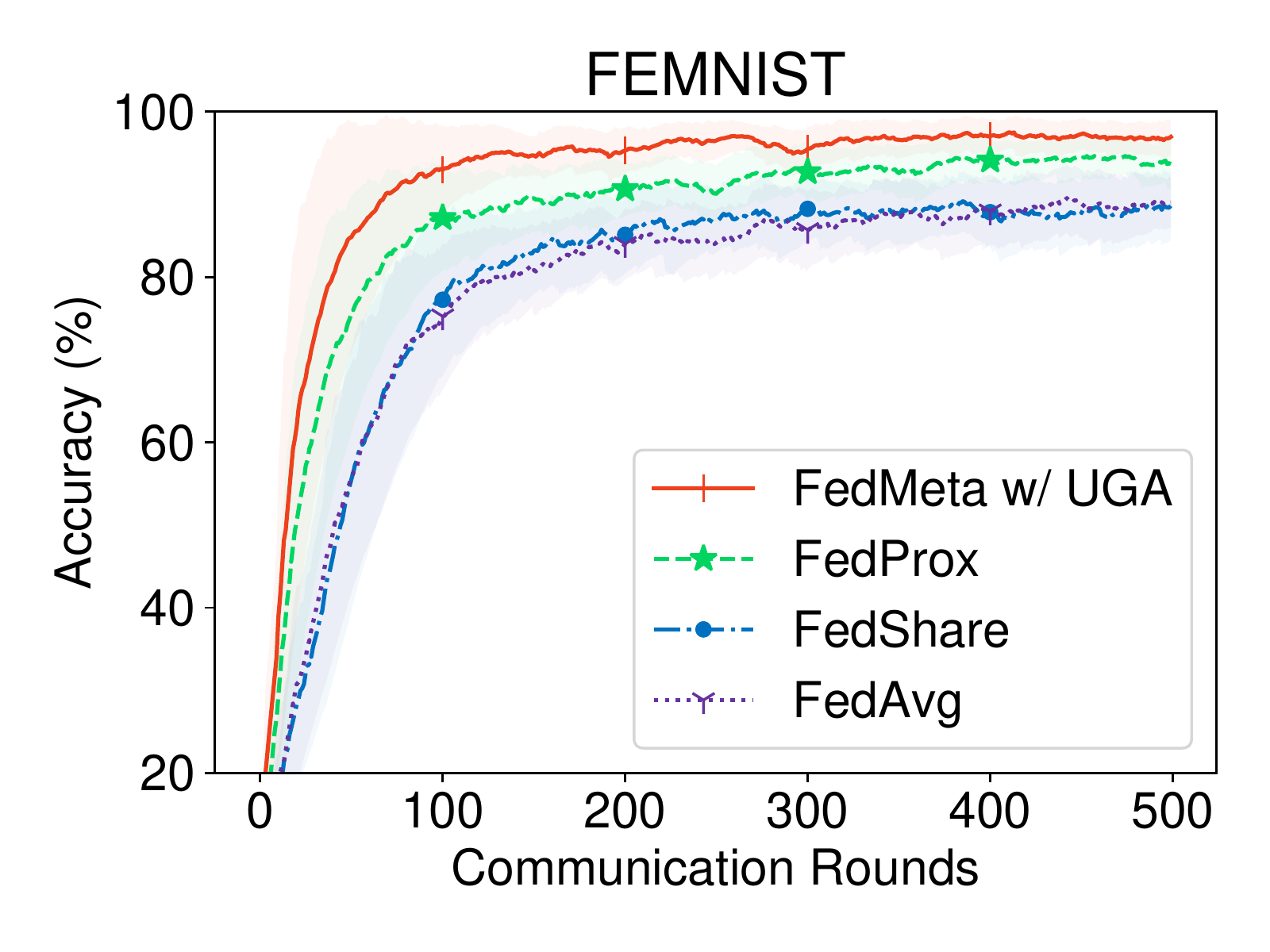}
        \includegraphics[width=\linewidth]{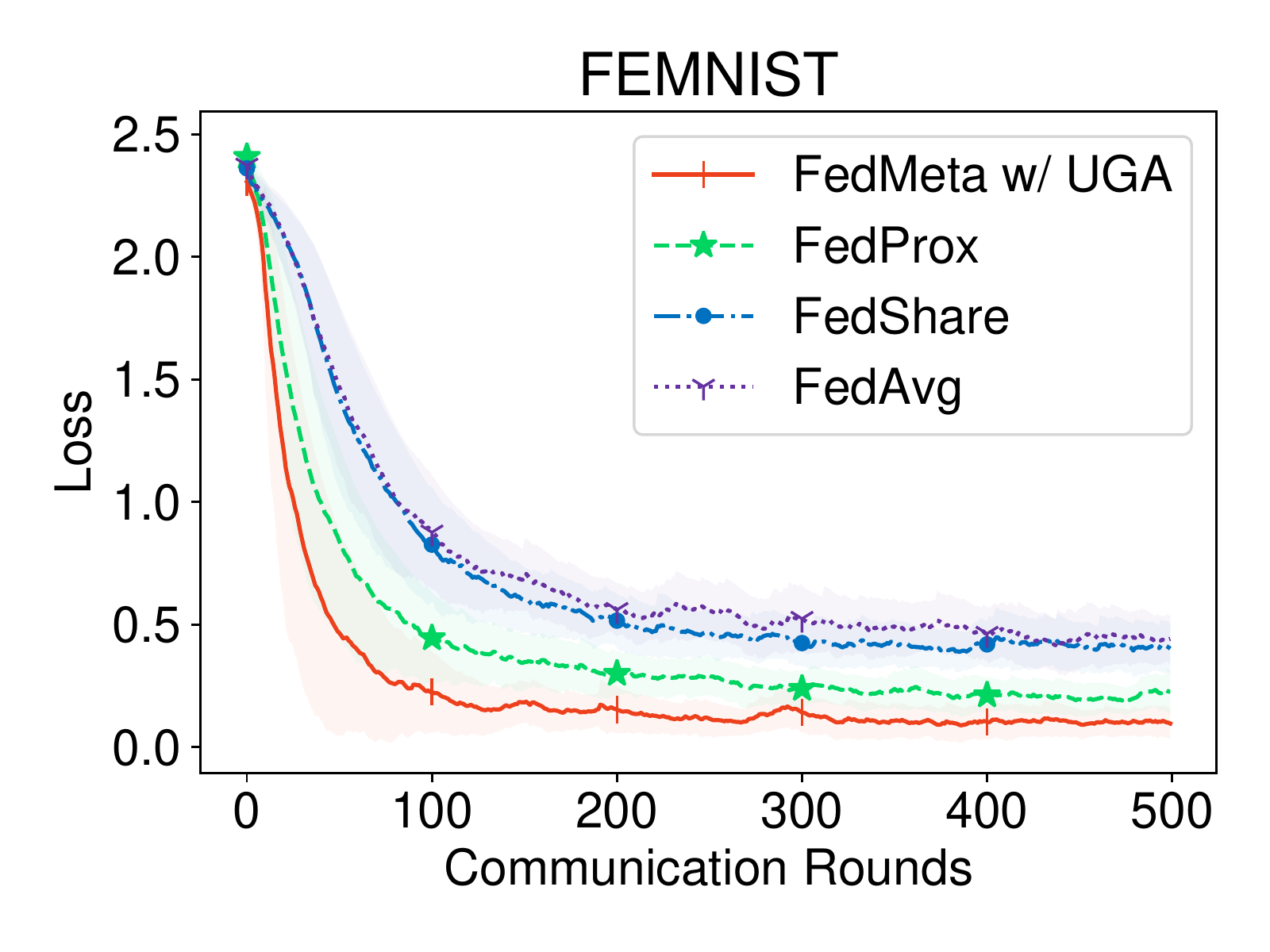}
        \caption{$E=5, B=64$}
        \label{fig:femnist_b}
    \end{subfigure}
    \begin{subfigure}{0.32\linewidth}
        \centering
        \includegraphics[width=\linewidth]{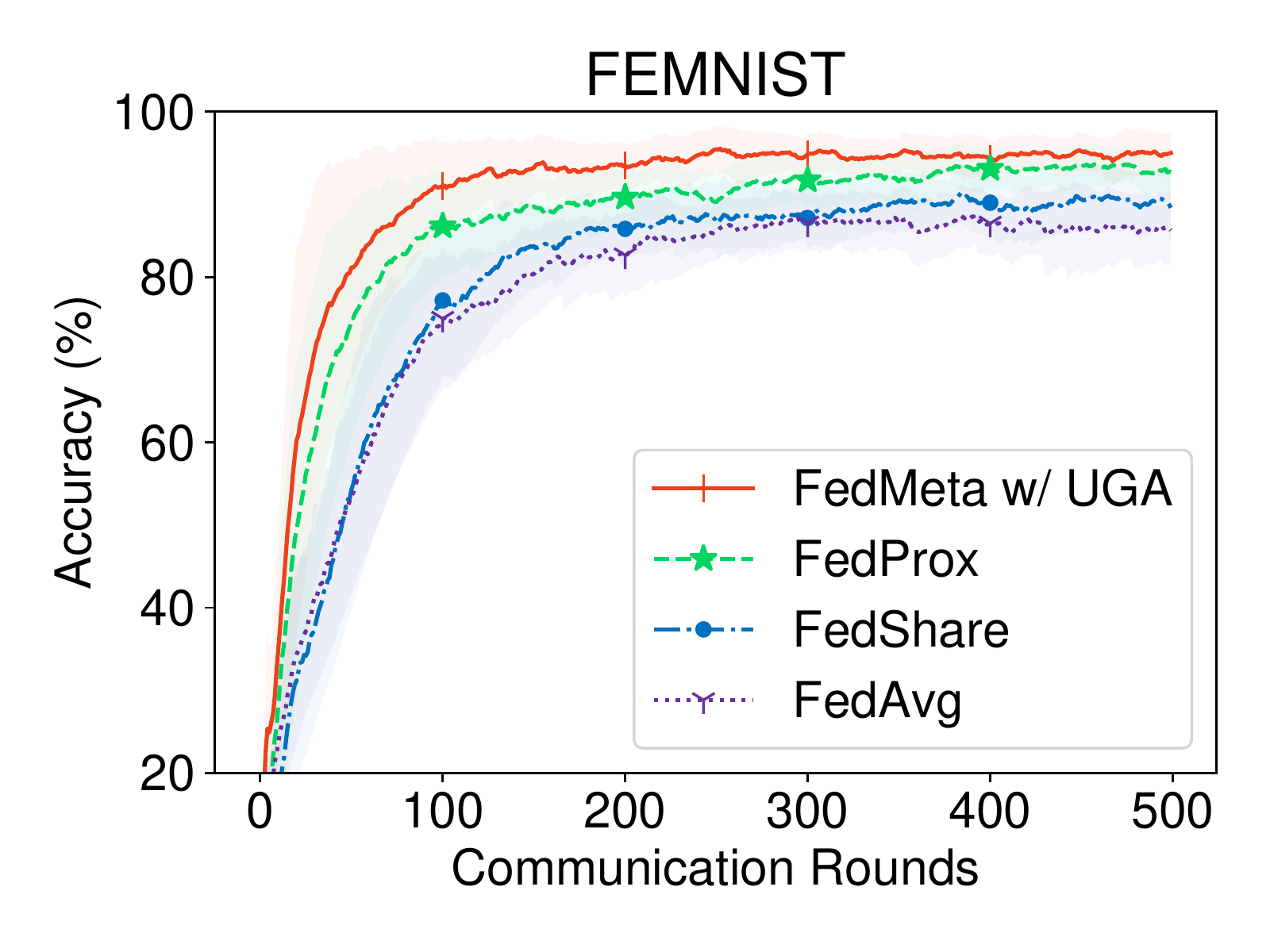}
        \includegraphics[width=\linewidth]{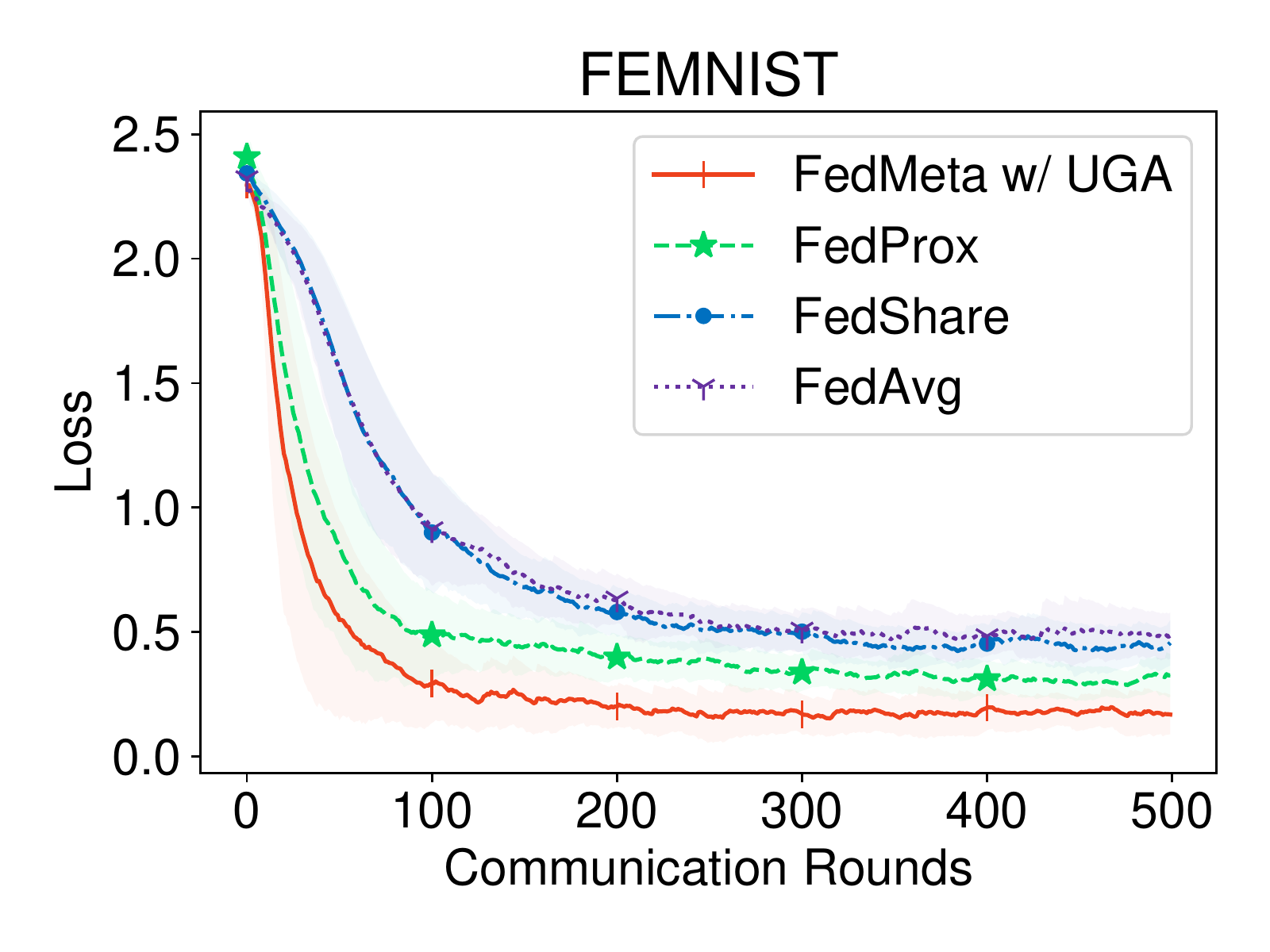}
        \caption{$E=5, B=128$}
        \label{fig:femnist_c}
    \end{subfigure}
    \caption{Test accuracy (upper row) and loss (lower row) over communication rounds of different methods with the CNN model on FEMNIST (non-IID): (a) $E=2, B=64$; (b) $E=5, B=64$; (c) $E=5, B=128$. (Better viewed in color)}
    \label{fig:femnist}
\end{figure*}

\begin{table}[t]
\caption{Number of communication rounds to reach accuracy milestones \& the convergence accuracy for all the methods on FEMNIST (non-IID) with $E=5, B=64$.}
\label{table:femnist}
\centering
\begin{threeparttable}
\begin{tabular}{@{}lcccc@{}}
\toprule
\multirow{2}{*}{Methods} & \multicolumn{3}{c}{Communication Rounds} & \multirow{2}{*}{\begin{tabular}[c]{@{}c@{}}Convergence\\ Accuracy\end{tabular}} \\ \cmidrule(lr){2-4}
                            & 70\%         & 80\%        & 90\%        &                                                                                 \\ \midrule
FedAvg                   & 68           & 111         & 437         & 90.22                                                                           \\
FedShare                 & 65           & 93          & 385         & 90.74                                                                           \\
FedProx                  & 32           & 50          & 144         & 95.27                                                                           \\ \midrule
FedMeta w/ UGA           & \textbf{21}  & \textbf{31} & \textbf{59} & \textbf{98.18}                                                                  \\ \bottomrule
\end{tabular}
\begin{tablenotes}
\item[*] Bold fonts indicate better performances, i.e., fewer communication rounds or higher accuracy.
\end{tablenotes}
\end{threeparttable}
\end{table}

\subsection{CNN Model on FEMNIST (non-IID)}
\label{sec:exp_femnist}

\subsubsection{Dataset}
FEMNIST is built by partitioning the data in Extended MNIST~\cite{cohen2017emnist} based on the writer of the digits.
It contains 3,383 writers with 341,873 training examples and 40,832 test examples in total.
We randomly select 100 writers and assign each writer (and its examples) to a client, making up 100 clients with typical non-IID data distributions.
Additionally, we randomly sample 1\% images as the sharing data\footnote{1\% is actually the average number of images on clients, which means $2\times$ images on clients if shared as a whole. To make a fair comparison with other methods, we further split them to the training client uniformly.} for FedShare and the meta training set for FedMeta.

\subsubsection{Model}
We use the same CNN architecture as FedAvg:
two 5$\times$5 convolution layers (the first with 32 channels while the second with 64, each followed by a ReLU activation and 2$\times$2 max pooling), a fully connected layer with 512 units (followed by a ReLU activation and random dropout), and a final softmax output layer.

\subsubsection{Hyper parameters}
$C=0.1$ (10 clients) is fixed for experiments on FEMNIST.
We use the SGD optimizer with the learning rate $\eta=0.002$ (with the linear scaling rule~\cite{goyal2017accurate} for different $B$) and a decay rate $=0.992$ per communication round on local clients.
We keep $\eta_g=0.002$ for UGA and $\eta_{meta}=\eta$ for FedMeta.
The coefficient for the proximal term in FedProx is set to $2\times10^{-4}$.

\subsubsection{Results}
The test accuracy and loss over communication rounds of \ours{} and the compared methods with different $E$ and $B$ are illustrated in Fig. \ref{fig:femnist}.
As shown in the figure, \ours outperforms FedProx, FedShare, and FedAvg in the convergence speed as well as the final rate of accuracy with a large margin.

When $E$ is small (Fig.~\ref{fig:femnist_a}), FedProx, FedShare, and FedAvg are slow in convergence due to the insufficient local training.
As a contrast, FedMeta w/ UGA achieves far better performance at much fewer communication and computation costs.
With the same $E$, improvements made by \ours{} are more significant when the steps of local updating increase (Fig.~\ref{fig:femnist_b} vs. \ref{fig:femnist_c}), which is identical to the conclusion in Section \ref{sec:cifar}.

Additionally, we summarize the number of communication rounds needed to reach certain accuracy milestones and the final convergence accuracy with $E=5, B=64$ for all the methods as Table~\ref{table:femnist}.
To reach the accuracy of 90\%, FedMeta w/ UGA needs only 13.50\%, 15.32\% and 40.97\% of the communication rounds compared to the vanilla FedAvg, FedShare, and FedProx respectively, which indicates a significant reduction in the communication costs.
Meanwhile, FedMeta w/ UGA achieves a final convergence accuracy of 98.18\%, far ahead of the performance achieved by the other three methods.


\subsection{GRU Model on Shakespeare (non-IID)}
\label{sec:exp_shake}
\subsubsection{Dataset}
Shakespeare is a dataset built from \textit{The Complete Works of William Shakespeare}, where each speaking role is considered as a unique client.
It contains 715 roles with 16,068 training examples (lines of words) and 2,356 test examples in total.
Similarly, we randomly select 100 roles as the clients, and for sampling the sharing data for FedShare or meta training data for FedMeta.

\subsubsection{Model}
We train a character\footnote{We always use \emph{character} to refer to a one-byte string and \emph{role} to refer to a speaking role in the play.}-level GRU language model, which predicts the next character after reading the previous ones in a line.
The input characters are first embedded into a 256-dimensional space, and then passed through a GRU layer with 1024 units and a softmax output layer.

\subsubsection{Hyper parameters}
$C=0.1$ (10 clients) is fixed for experiments on Shakespeare.
We use the SGD optimizer with the learning rate $\eta=0.1(=\eta_{meta}\text{ for FedMeta})$ and a decay rate $=0.992$ per communication round on local clients.

\subsubsection{Results}
\begin{figure*}[t]
    \centering
    \begin{subfigure}{0.32\linewidth}
        \centering
        \includegraphics[width=\linewidth]{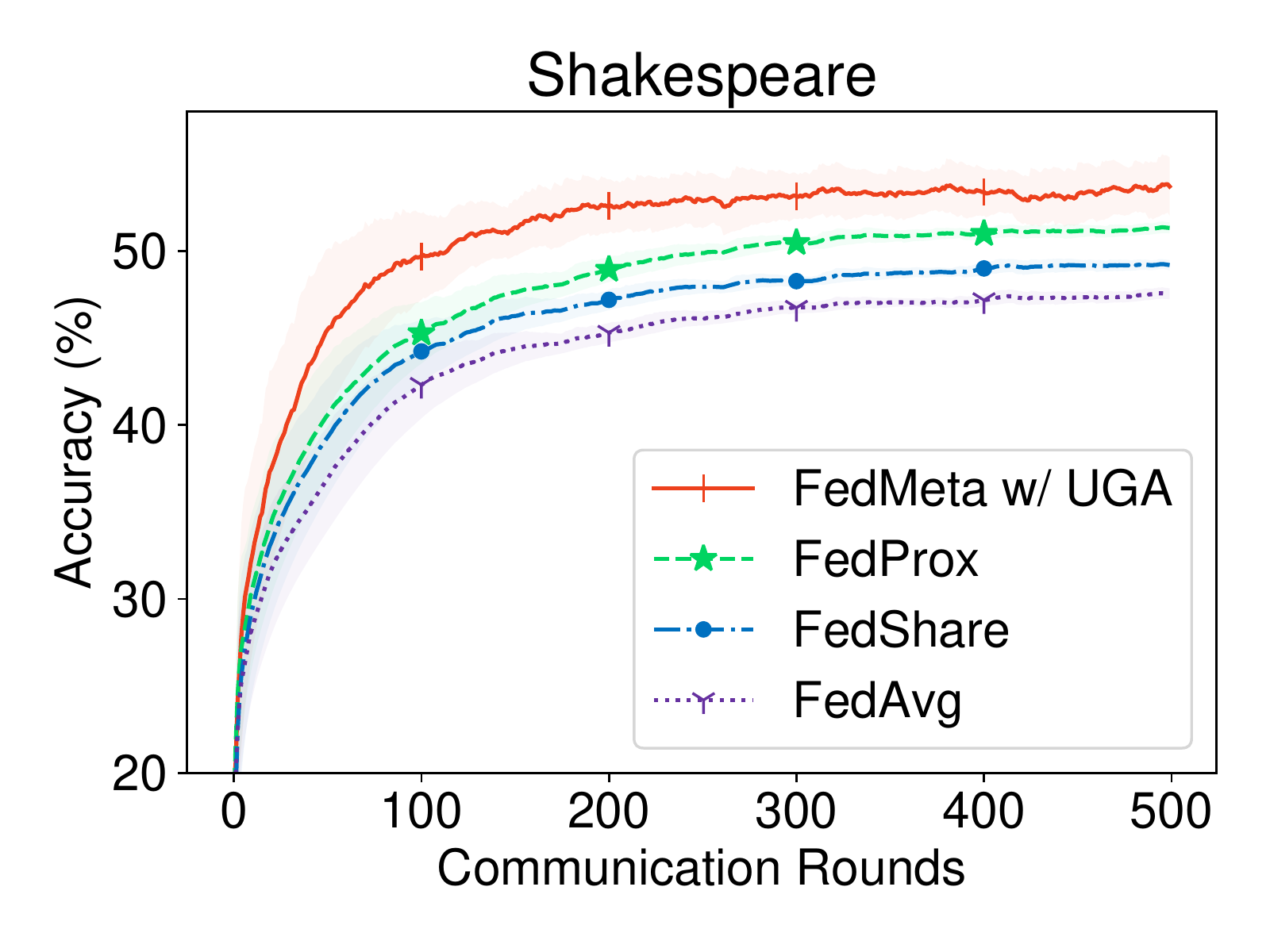}
        \caption{}
        \label{fig:shake_a}
    \end{subfigure}
    \begin{subfigure}{0.32\linewidth}
        \centering
        \includegraphics[width=\linewidth]{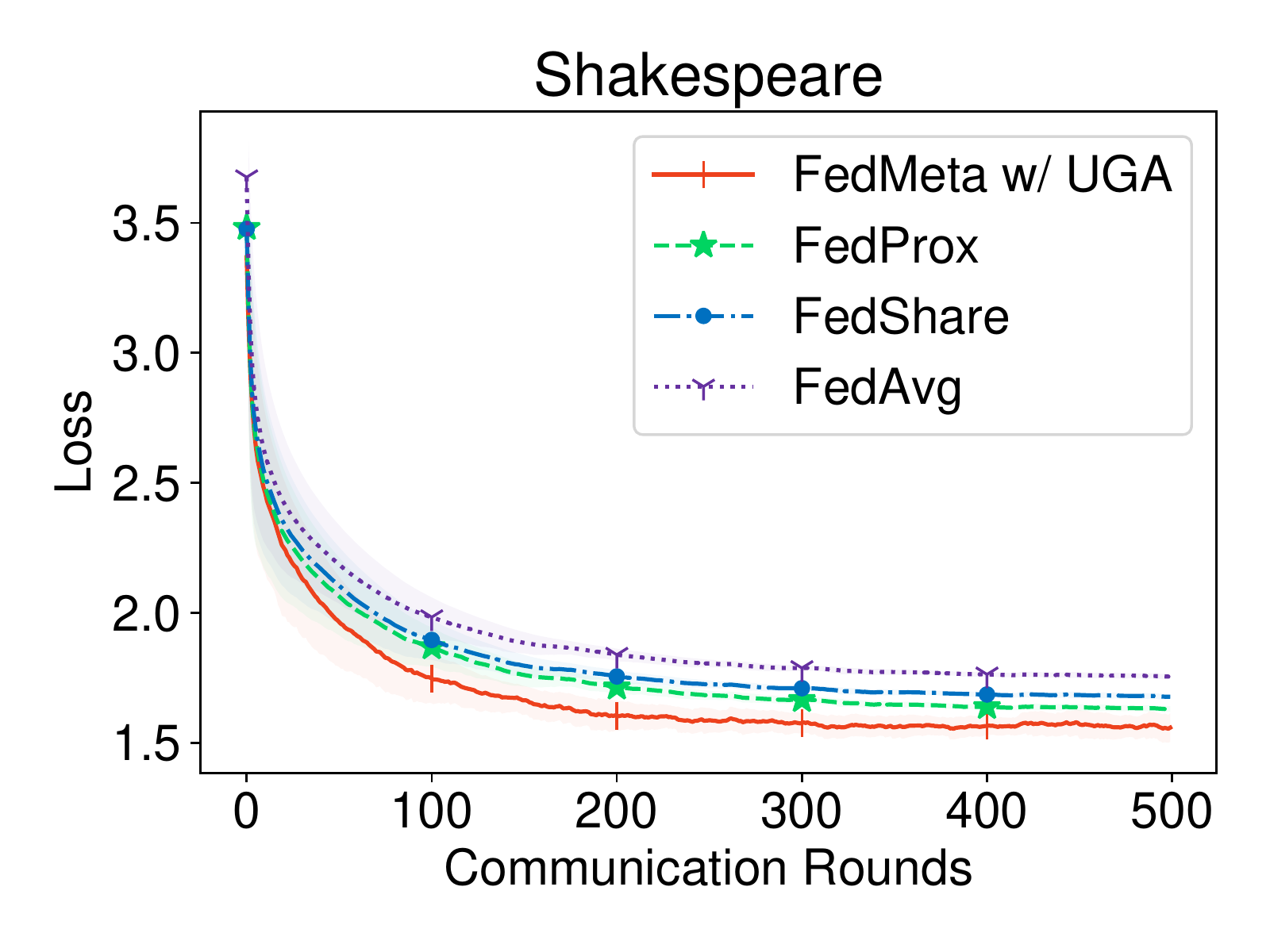}
        \caption{}
        \label{fig:shake_b}
    \end{subfigure}
    \begin{subfigure}{0.32\linewidth}
        \centering
        \includegraphics[width=\linewidth]{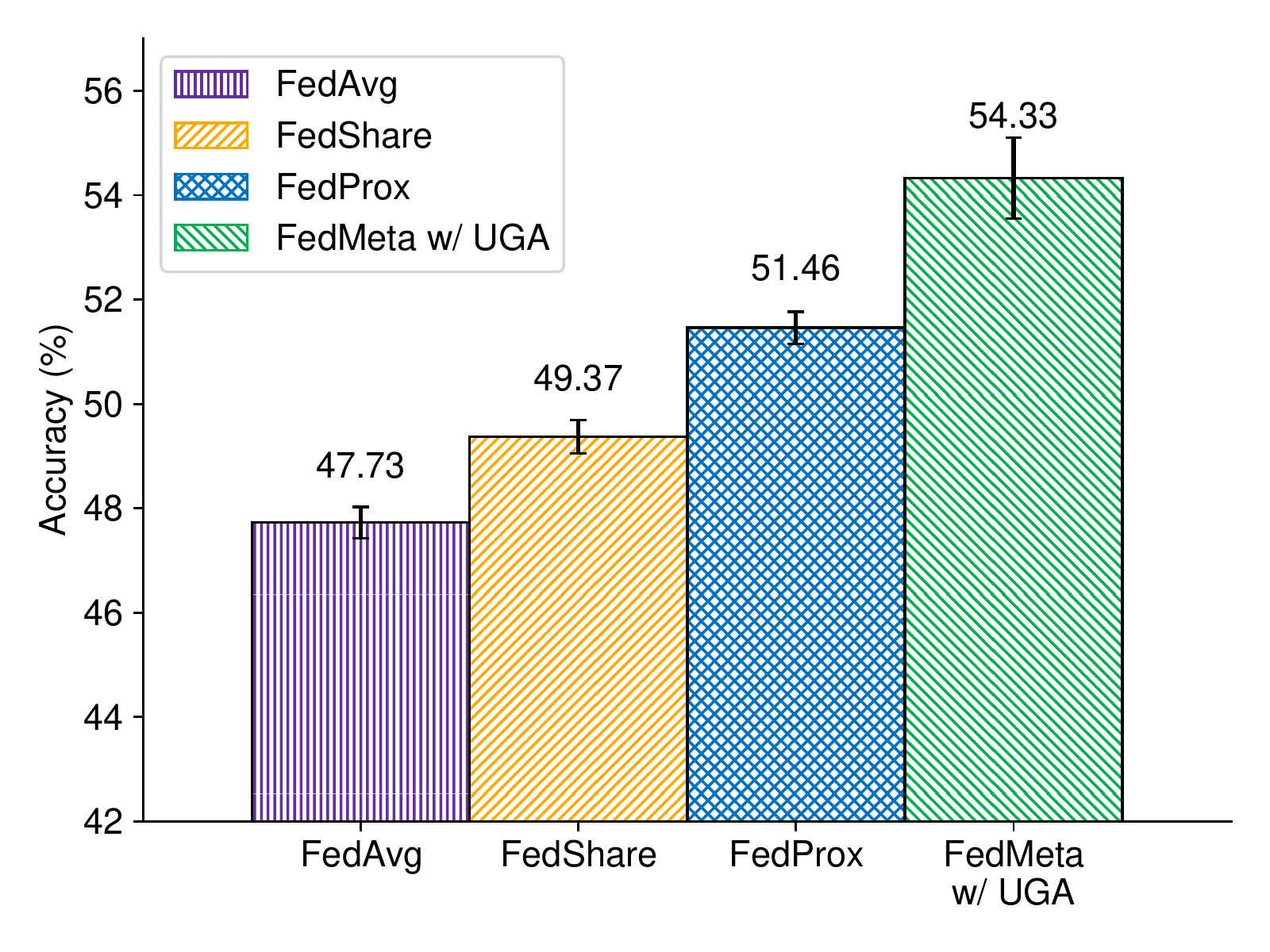}
        \caption{}
        \label{fig:shake_c}
    \end{subfigure}
    \caption{Experimental results of different methods with the GRU model on Shakespeare (non-IID): (a) Accuracy and (b) Loss over communication rounds; (c) the final convergence accuracy. (Better viewed in color)}
    \label{fig:shake}
\end{figure*}

The experimental results with the GRU model on Shakespeare is illustrated in Fig.~\ref{fig:shake}.
Similar to the conclusions in the above two experiments, the proposed \ours{} achieves higher accuracy than the compared methods, with increases in convergence accuracy by 13.83\%, 10.05\%, and 5.58\% compared to FedAvg, FedShare and FedProx respectively, as shown in Fig. \ref{fig:shake_c}.
The proposed \ours{} is also faster in convergence.
FedAvg, FedShare, and FedProx spend 319, 182, and 122 rounds respectively to achieve 47\% accuracy.
As a contrast, \ours{} needs only 49 communication rounds to reach a competitive performance.
On Shakespeare, FedShare slightly improve the performance of FedAvg, but still underperforms FedProx and \ours{}.
This may be due to that the data distributions are extremely non-IID on Shakespeare and the sampled sharing dataset is not representative enough.
For examples, the sharing data sampled here contains sentences from 100 roles, which would take negative effects on a certain client if the role had a rather different speaking habits from others.

GRU (and other recurrent neural network (RNN)) models usually require data to pass through the same network architecture far times than CNNs, which results in more serious gradient biases for FedAvg.
UGA alleviates the problem but suffers from the instabilities brought by back propagating derivatives through the same architecture multiple times, which accounts for the large variances in the final convergence accuracies of FedMeta w/ UGA as shown by the variance bar in Fig.~\ref{fig:shake_c}.

\begin{figure}[t]
    \centering
    \includegraphics[width=0.7\linewidth]{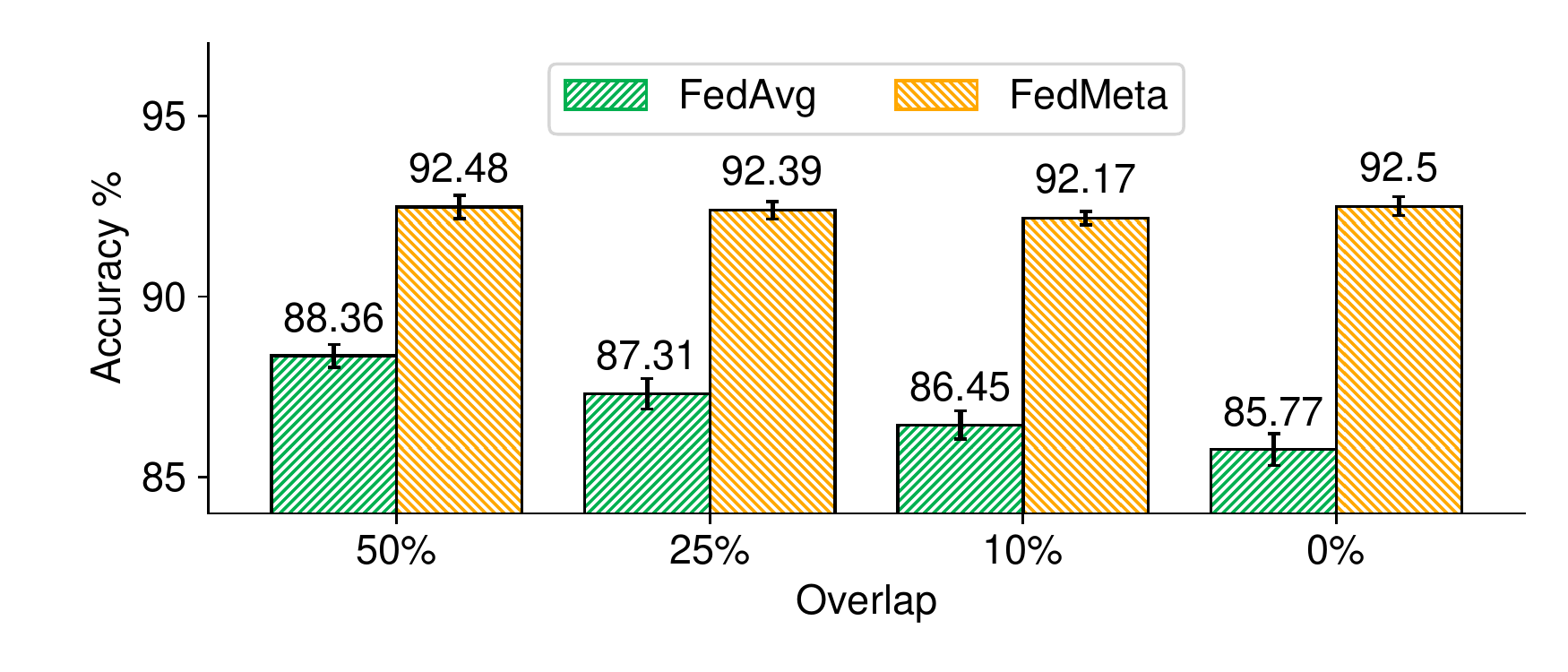}
    \caption{Test accuracy with different overlap rates between $\mathcal{D}_{meta}$ and $\mathcal{D}$ on FEMNIST. (Better viewed in color)}
    \label{fig:overlap}
\end{figure}

\subsection{Controllable $\mathcal{D}_{meta}$}

In Section~\ref{sec:method_2}, we propose that $\mathcal{D}_{meta}$ offers a way to control the behavior of federated models.
Here we simulate this situation by constructing a $\mathcal{D}_{meta}$ that has partial or no overlap writers in FEMNIST with $\mathcal{D}$.

\subsubsection{Setting}
The dataset partition for $\mathcal{D}$, model and main hyper parameters follow those in Section \ref{sec:exp_femnist}.
Additionally, we sample another 100 writers, together with their examples, to form the auxiliary dataset.
Then we construct $\mathcal{D}_{meta}$ by sampling examples from the mixture of $\mathcal{D}$ and the auxiliary dataset with a certain proportion.
For example, \emph{25\% overlap} means there are 25\% writers in $\mathcal{D}_{meat}$ are also included in $\mathcal{D}$, while the other 75\% writers comes from the auxiliary dataset.
In practice, we first select writers from $\mathcal{D}$ and the auxiliary dataset according to a certain proportion, and then sample 1\% examples to form the $\mathcal{D}_{meta}$.

\subsubsection{Results}
The test accuracies of FedAvg and FedMeta (without UGA) on FEMNIST with various overlap rates between $\mathcal{D}_{meta}$ and $\mathcal{D}$ are shown in Fig.~\ref{fig:overlap}.
The accuracy of FedAvg drops significantly, from 88.36\% to 85.77\%, as the divergence between $\mathcal{D}_{meta}$ and $\mathcal{D}$ increases.
It is worth noting that when $\mathcal{D}_{meta}$ is sampled from $\mathcal{D}$, i.e., 100\% overlap rate, the accuracy is 90.31\%.
It shows that if $\mathcal{D}_{meta}$ had a large difference from $\mathcal{D}$, the accuracy of FedAvg could drop as much as 5 percentage points.
As a contrast, under the guidance of the meta training set, FedMeta keep its performance around 92\% all the time.

\subsection{Ablation Study on UGA and FedMeta}
\label{sec:ab}

Since we propose two improvements, i.e., UGA (Section \ref{sec:method_1}) and FedMeta (Section \ref{sec:method_2}), to tackle the deficiencies in FedAvg, we conduct ablation studies here to take a closer look at their contributions.

\subsubsection{Ablation Experiments on FEMNIST (non-IID)}
\begin{figure*}[t]
    \centering
    \begin{subfigure}{0.32\linewidth}
        \centering
        \includegraphics[width=\linewidth]{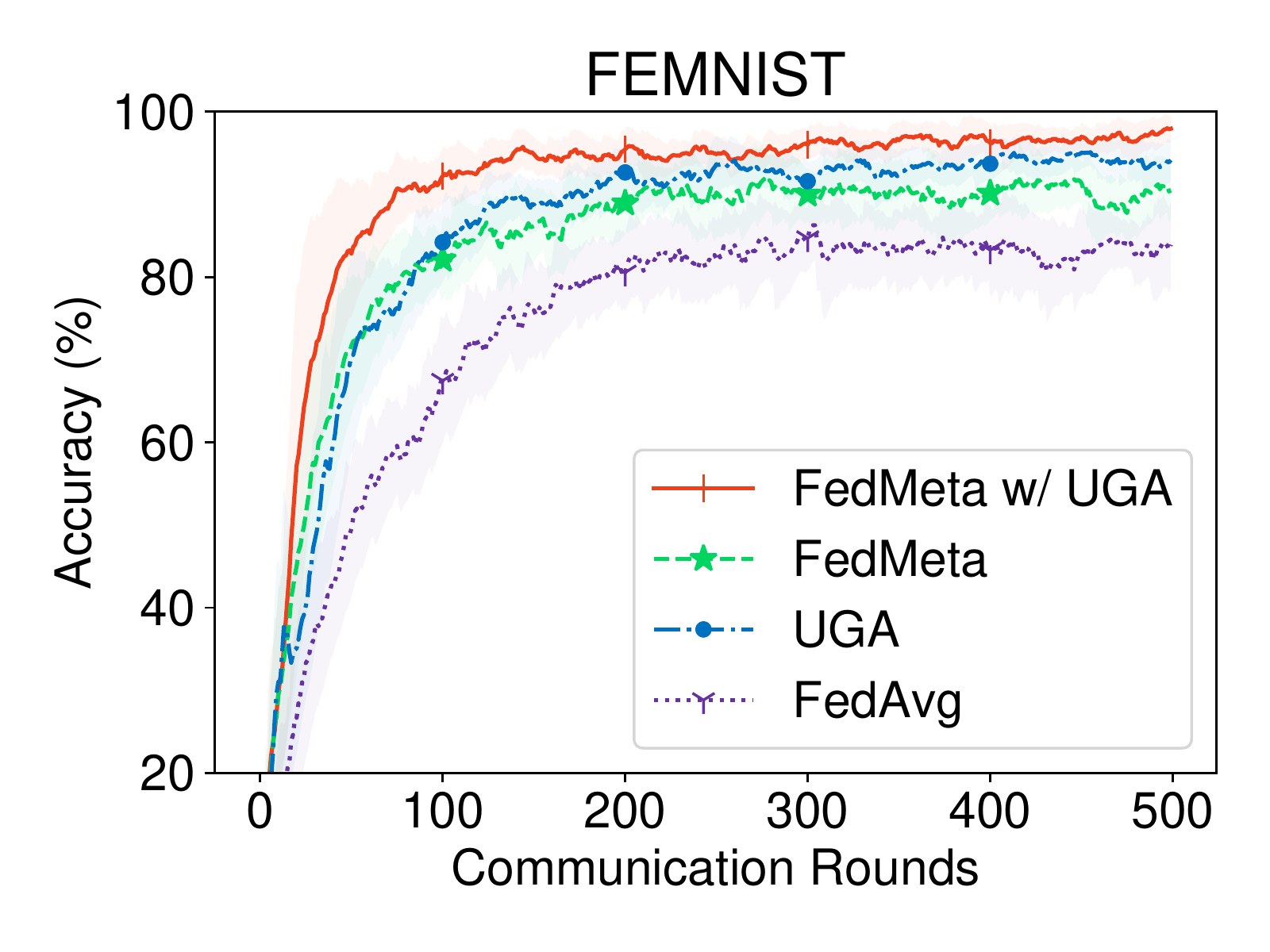}
        \includegraphics[width=\linewidth]{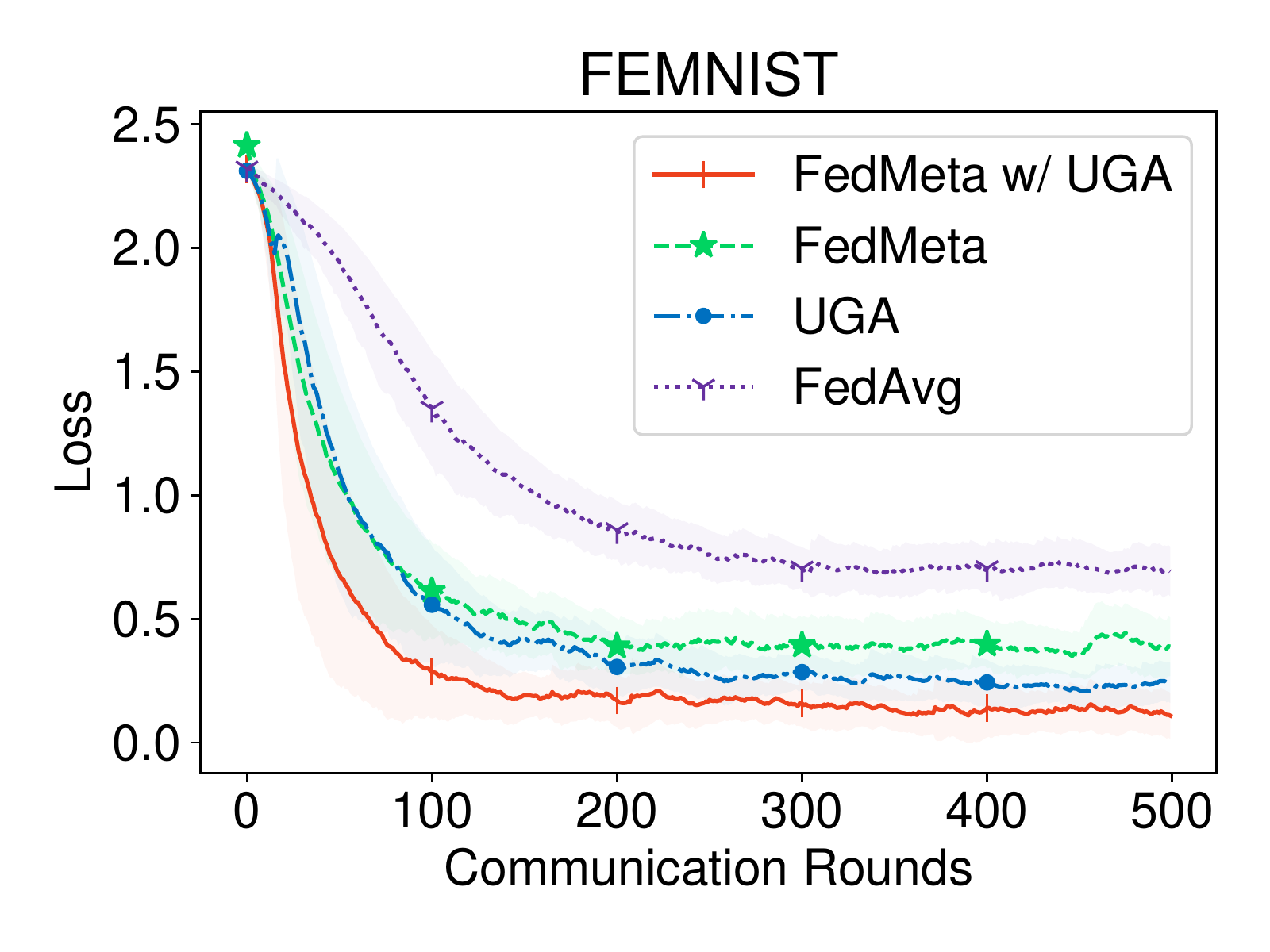}
        \caption{$E=2, B=64$}
        \label{fig:ab_femnist_a}
    \end{subfigure}
    \begin{subfigure}{0.32\linewidth}
        \centering
        \includegraphics[width=\linewidth]{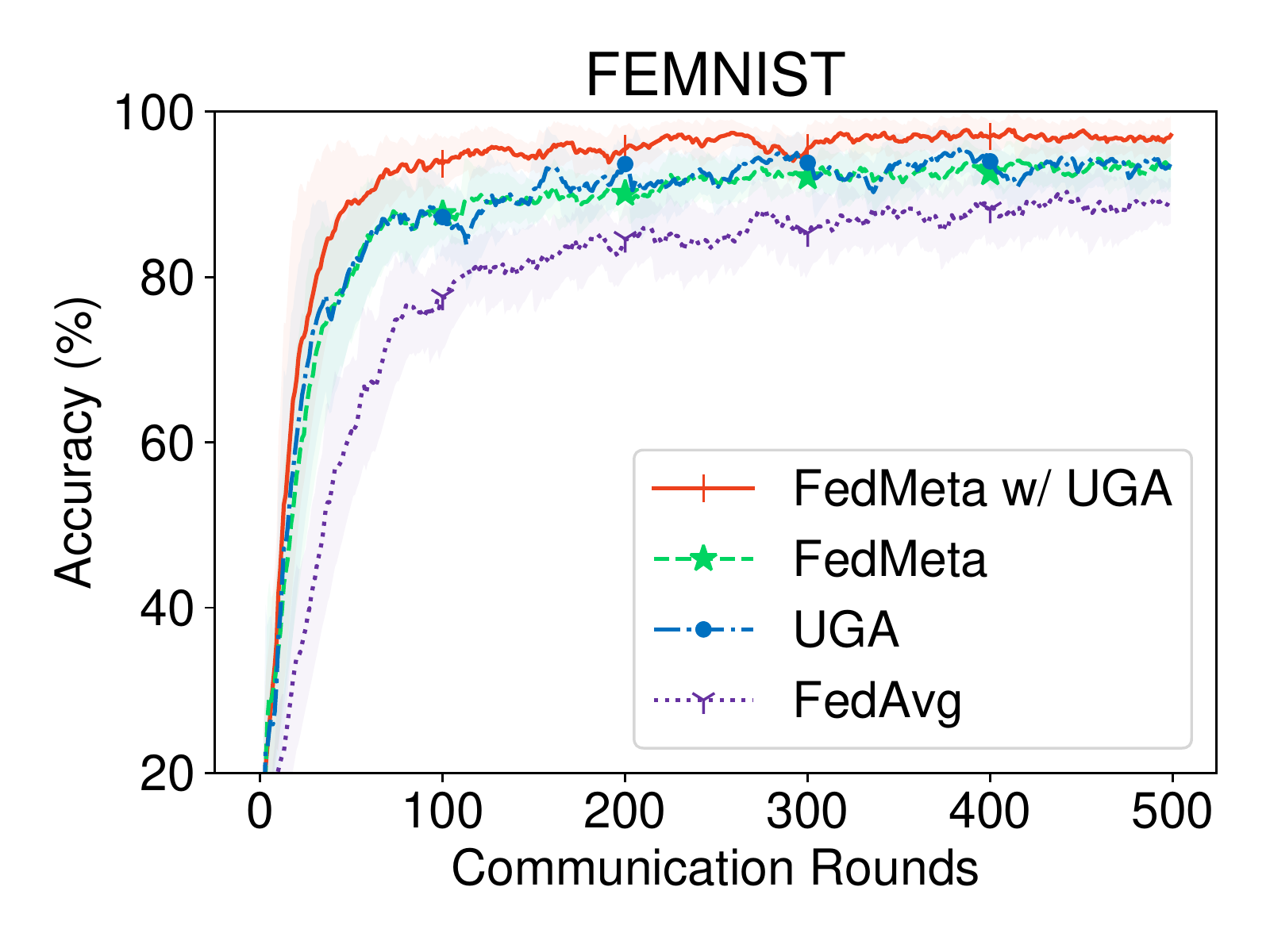}
        \includegraphics[width=\linewidth]{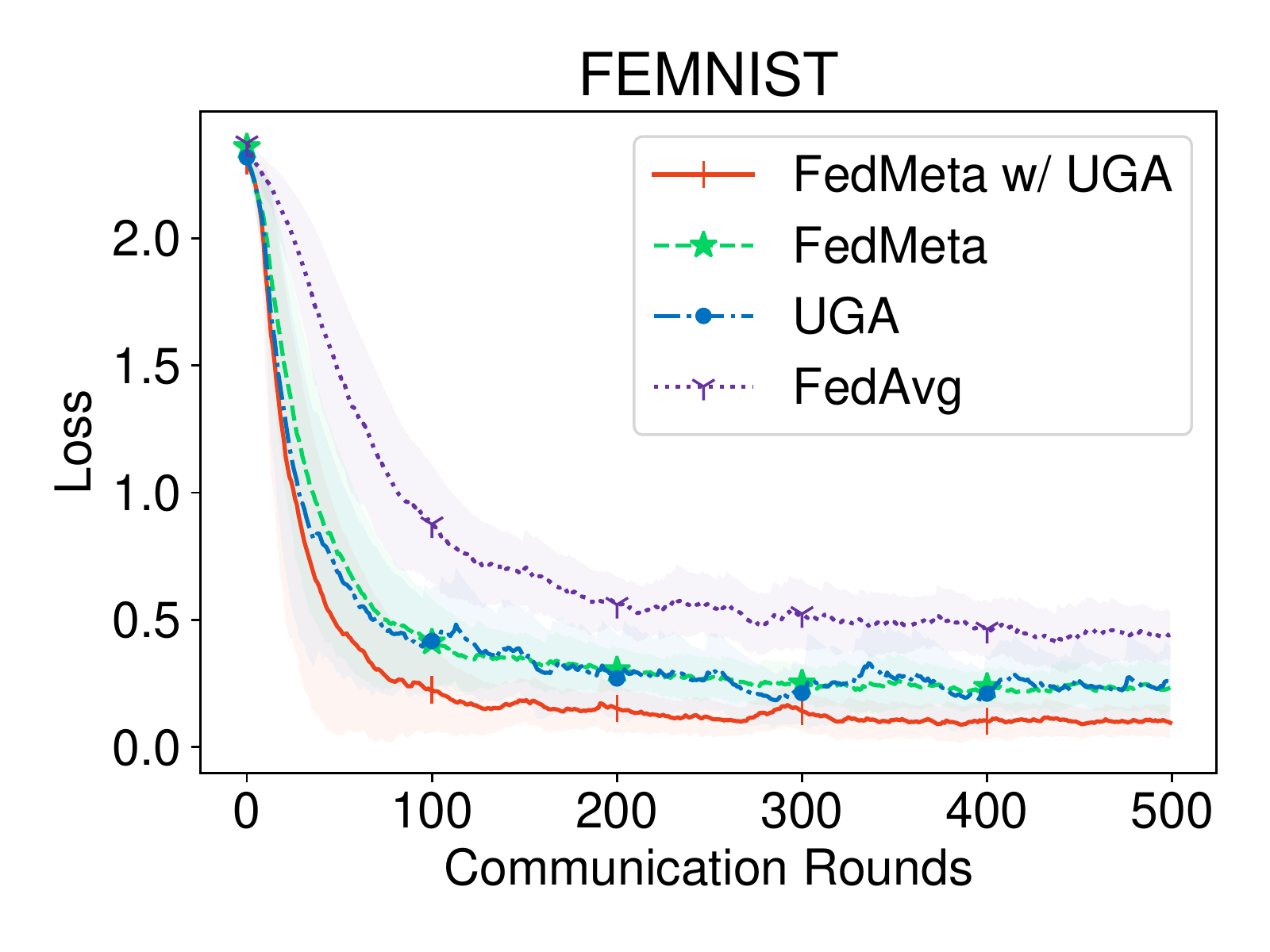}
        \caption{$E=5, B=64$}
        \label{fig:ab_femnist_b}
    \end{subfigure}
    \begin{subfigure}{0.32\linewidth}
        \centering
        \includegraphics[width=\linewidth]{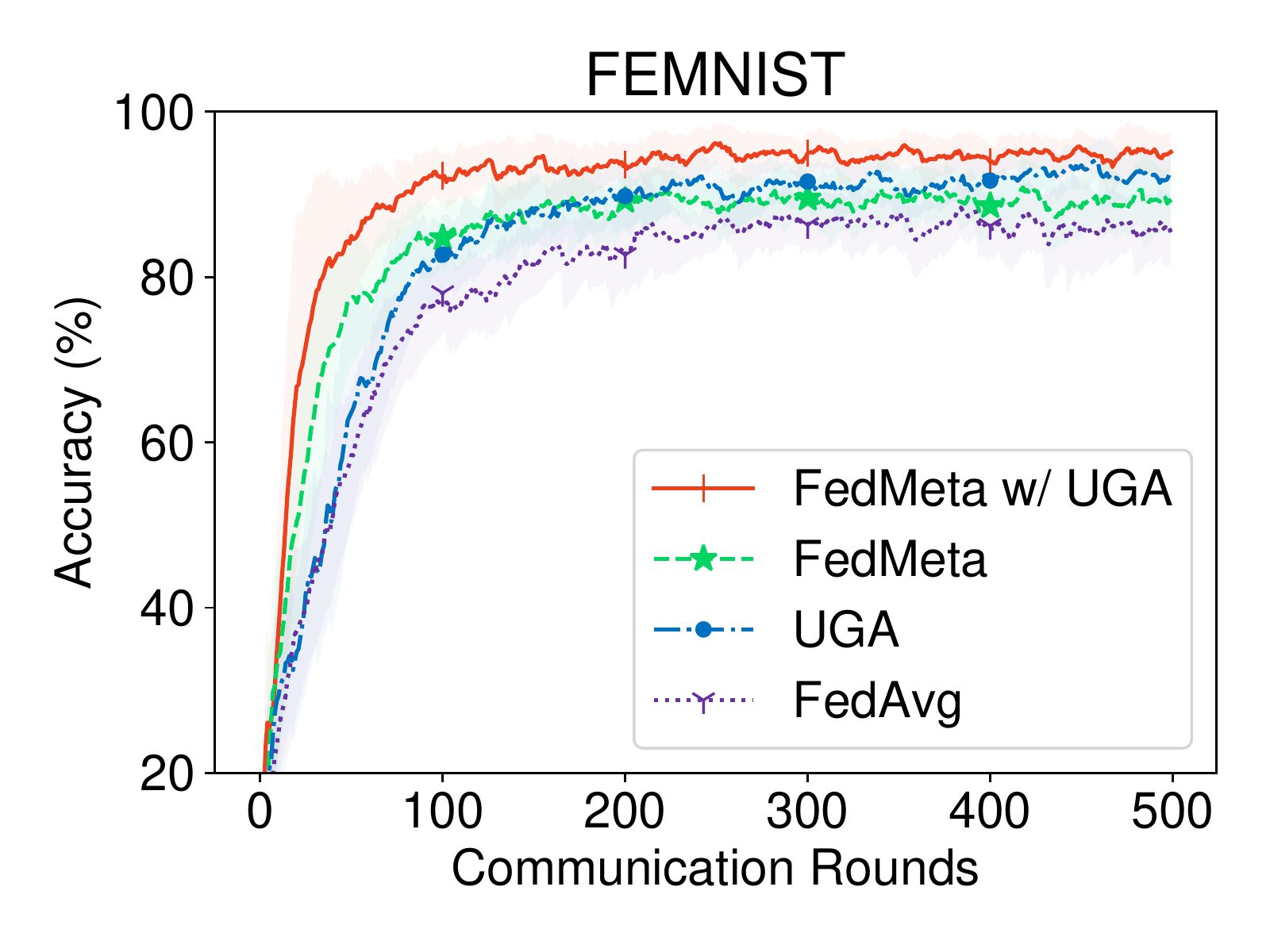}
        \includegraphics[width=\linewidth]{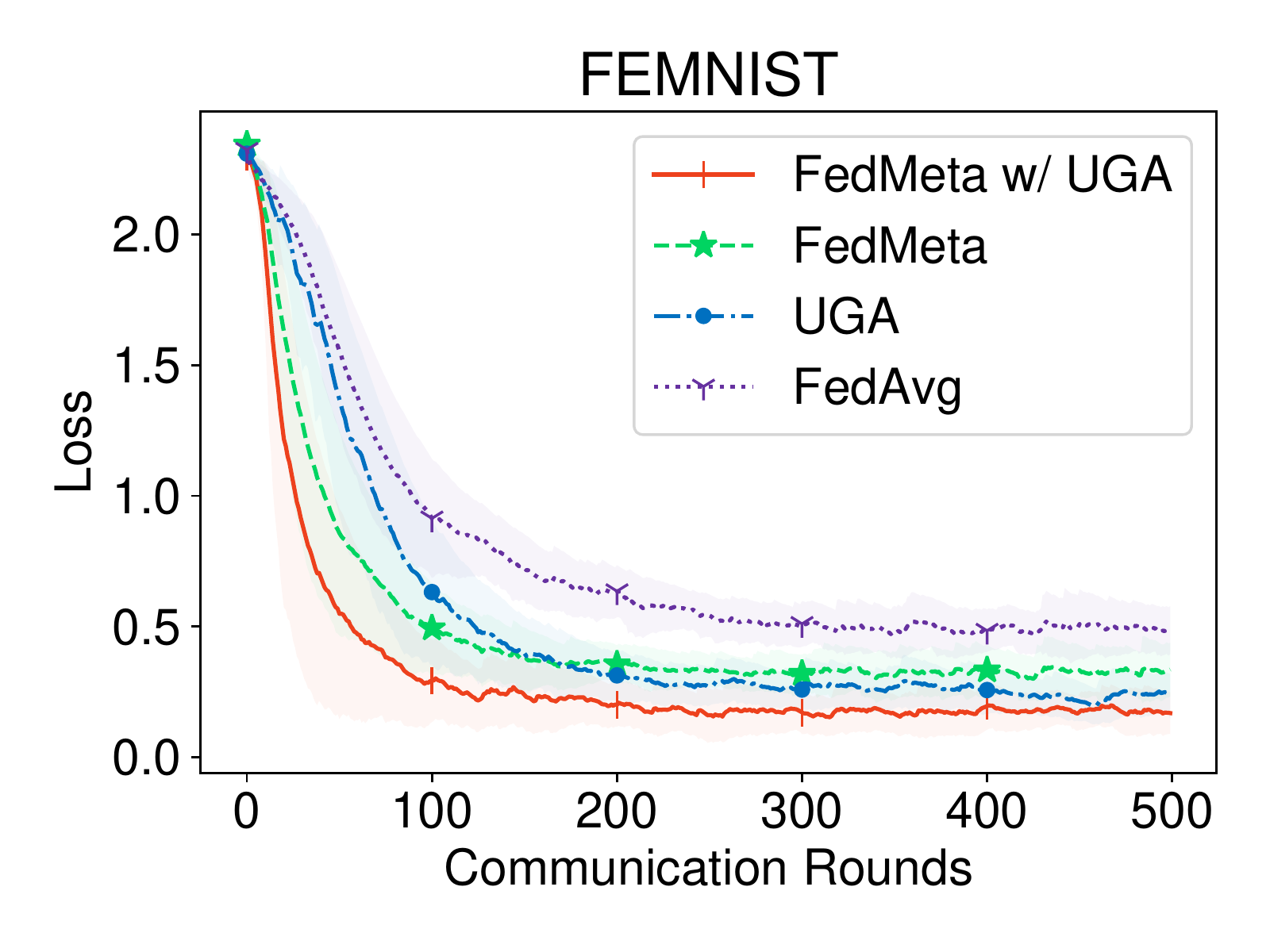}
        \caption{$E=5, B=128$}
        \label{fig:ab_femnist_c}
    \end{subfigure}
    \caption{Ablation study results of UGA and FedMeta on FEMNIST (non-IID): test accuracy (upper row) and loss (lower row): (a) $E=2, B=64$; (b) $E=5, B=64$; (c) $E=5, B=128$. (Better viewed in color)}
    \label{fig:ab_femnist}
\end{figure*}

We illustrate the test accuracy and loss over communication rounds of UGA and FedMeta in Fig. \ref{fig:ab_femnist}.
To make a straightforward comparison, we also plot the learning curves of FedAvg as baselines, and \ours{} as the upper bounds.

Considering the performances of FedMeta and UGA separately, they both show some improvements over FedShare and FedAvg.
Generally, UGA achieves better performances than FedMeta, especially when there are not many steps of local updating (Fig.~\ref{fig:ab_femnist_a} vs. \ref{fig:ab_femnist_c}).
When the steps of local updating increase, UGA reaches comparable accuracy but is less stable than FedMeta with larger jitters in accuracy and loss (as shown in Fig.~\ref{fig:ab_femnist_b}).
This may be caused by back propagating derivatives through the unfolded loop consisting of the same network multiple times in UGA, as pointed out by~\cite{antoniou2018train}.

Similar to that in Section \ref{sec:exp_femnist}, we further summarize the number of communication rounds to reach accuracy milestones and the convergence accuracy of UGA and FedMeta on FEMNIST with $E=5, B=64$ in Table \ref{table:ab_femnist}.
UGA and FedMeta both outperform FedAvg in convergence speed as well as the final convergence accuracy.
Concretely, UGA and FedMeta achieve the accuracy of 70\% and 80\% with very close communication rounds, fewer than half of FedAvg, while UGA reaches slightly higher convergence accuracy.
It shows that UGA performs similarly to FedMeta at the beginning but can further improve the federated model as the training process goes on.

\begin{table}[t]
\caption{Number of communication rounds to reach accuracy milestones \& the convergence accuracy of UGA and FedMeta separately on FEMNIST (non-IID). ($E=5, B=64$)}
\label{table:ab_femnist}
\centering
\begin{threeparttable}
\begin{tabular}{@{}lcccc@{}}
\toprule
\multirow{2}{*}{Methods} & \multicolumn{3}{c}{Communication Rounds} & \multirow{2}{*}{\begin{tabular}[c]{@{}c@{}}Convergence\\ Accuracy\end{tabular}} \\ \cmidrule(lr){2-4}
                            & 70\%         & 80\%        & 90\%        &                                                                                 \\ \midrule
FedAvg                   & 68           & 111         & 437         & 90.22                                                                           \\ \midrule
UGA                      & 27           & 48          & 137         & 95.87                                                                           \\
FedMeta                  & 30           & 49          & 155         & 94.98                                                                           \\ \midrule
FedMeta w/ UGA           & \textbf{21}  & \textbf{31} & \textbf{59} & \textbf{98.18}                                                                  \\ \bottomrule
\end{tabular}
\begin{tablenotes}
\item[*] Bold fonts indicate better performances, i.e., fewer communication rounds or higher accuracy.
\end{tablenotes}
\end{threeparttable}
\end{table}

\begin{figure*}[t]
    \centering
    \begin{subfigure}{0.32\linewidth}
        \centering
        \includegraphics[width=\linewidth]{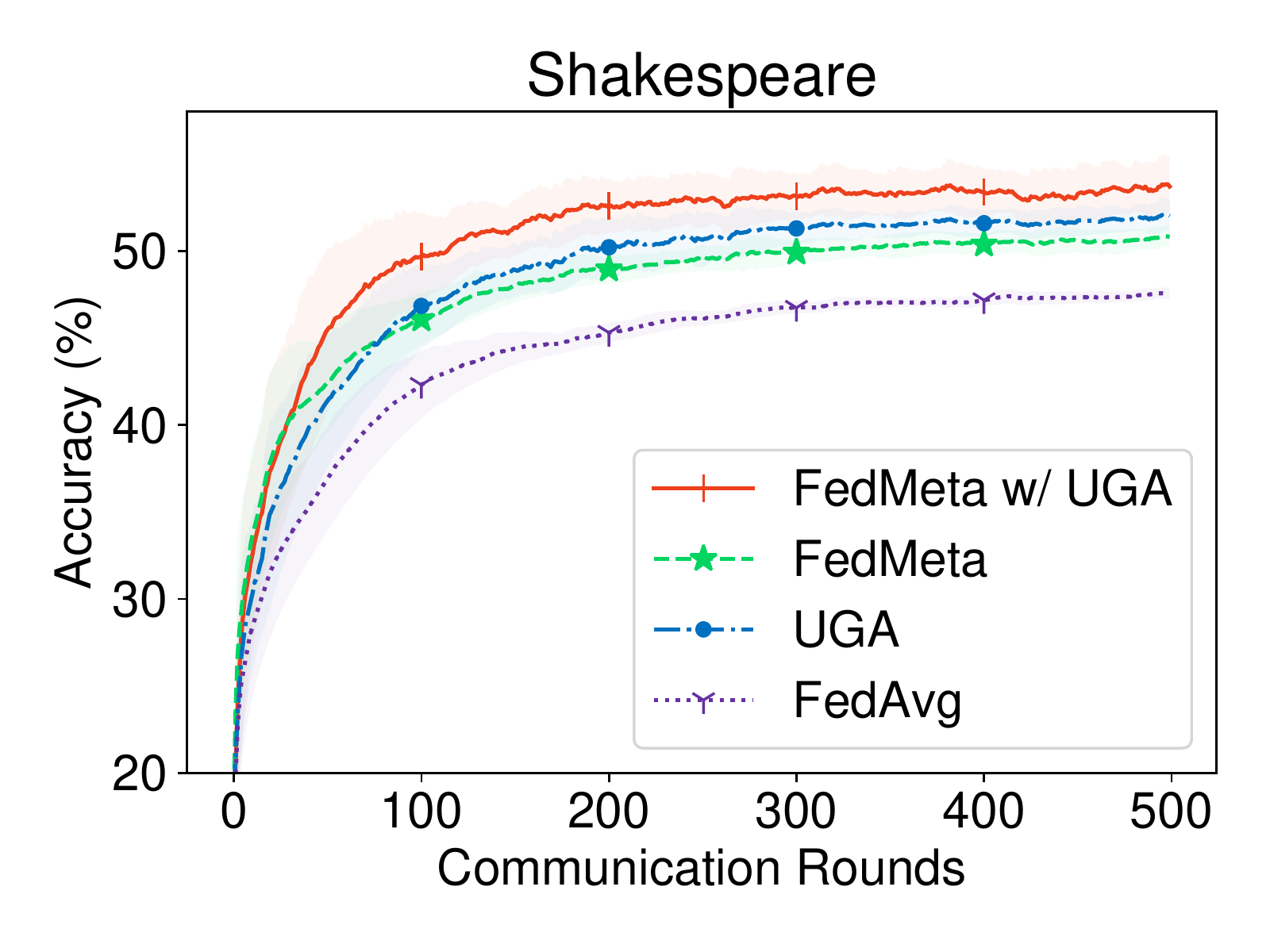}
        \caption{}
        \label{fig:ab_shake_a}
    \end{subfigure}
    \begin{subfigure}{0.32\linewidth}
        \centering
        \includegraphics[width=\linewidth]{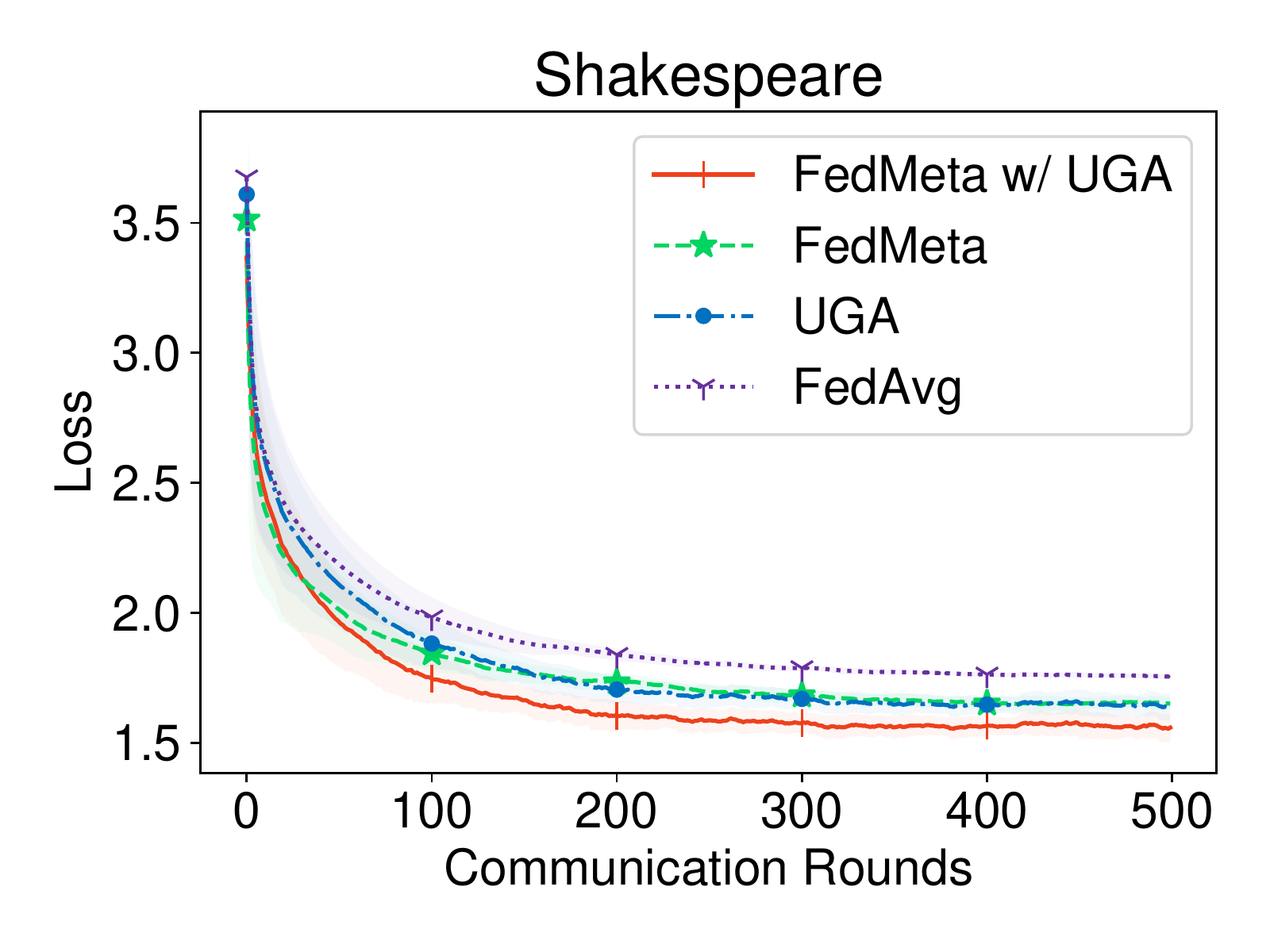}
        \caption{}
        \label{fig:ab_shake_b}
    \end{subfigure}
    \begin{subfigure}{0.32\linewidth}
        \centering
        \includegraphics[width=\linewidth]{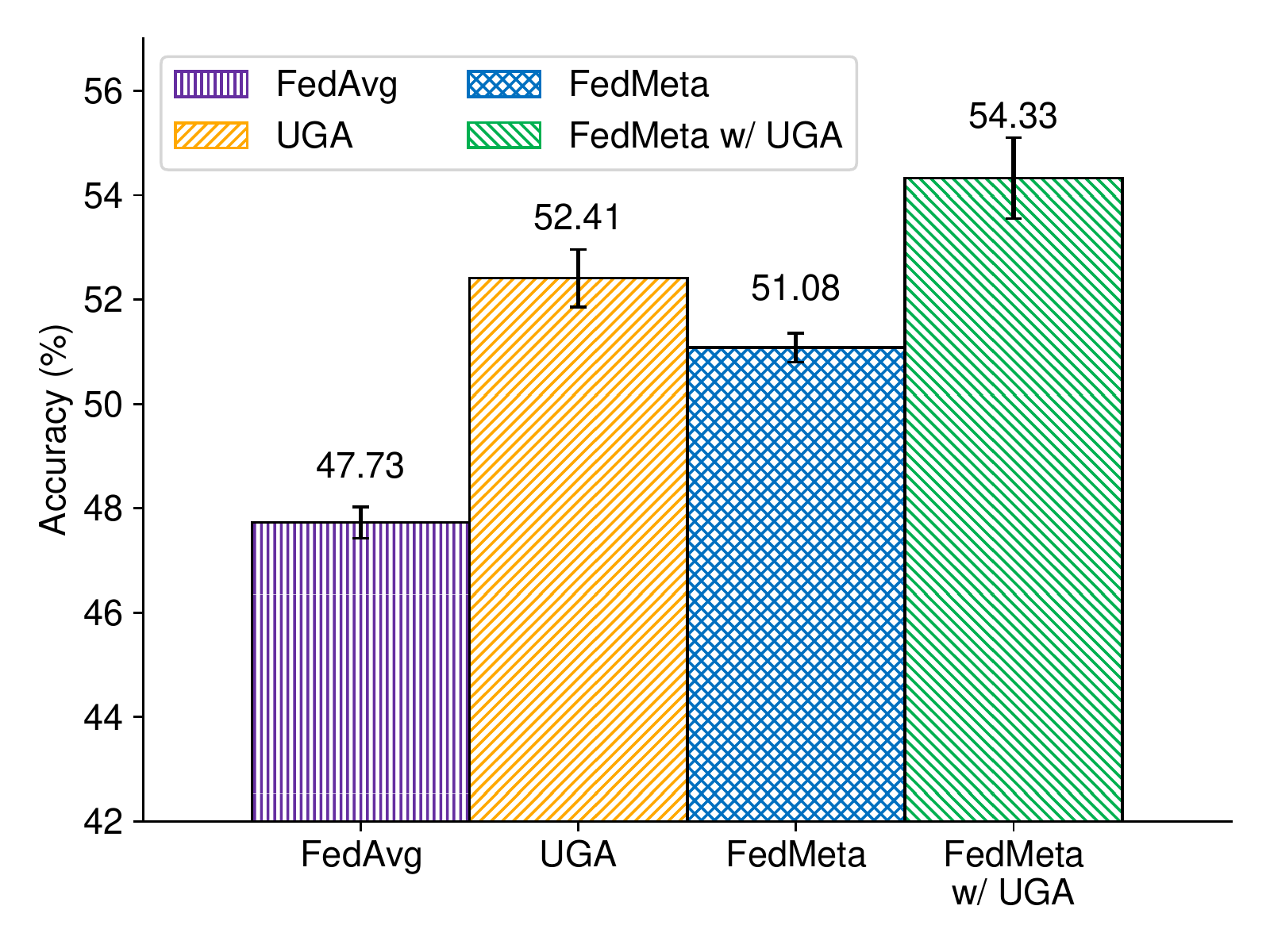}
        \caption{}
        \label{fig:ab_shake_c}
    \end{subfigure}
    \caption{Ablation study results of different methods with the GRU model on Shakespeare (non-IID): (a) Accuracy and (b) Loss over communication rounds; (c) the final convergence accuracy. (Better viewed in color)}
    \label{fig:ab_shake}
\end{figure*}

\subsubsection{Ablation Experiments on Shakespeare (non-IID)}

We also conduct ablation experiments on Shakespeare (non-IID) to study how the proposed methods UGA and FedMeta perform with RNN models.
The basic settings for model architecture and hyper parameters are identical to those in Section \ref{sec:exp_shake}.

The experimental results are illustrated in Fig. \ref{fig:ab_shake}.
Similarly, we plot the performance of FedAvg and \ours{} for comparison.
As we can see, both UGA and FedMeta outperform the baseline method FedAvg in convergence speed as well as the convergence accuracy, but underperform \ours{}.
Concretely, UGA performs slightly better than FedMeta and achieves 52.41\% accuracy, more than one percentage point higher than FedMeta.
The reason have been explained in Section \ref{sec:exp_shake}:
RNN models usually require data to pass through the same network architecture far many times, which results in serious gradient biases in FedAvg.
UGA greatly alleviates this problem and thus achieves better performance.

%% file: 5-others.tex
\section{Related Work}
\label{sec:related_work}

Our work is mainly related to the model- and task-agnostic performance improvements over FedAvg.
Besides, the proposed UGA with the keep-trace gradient descent and FedMeta w/ UGA are also technically related to the model-agnostic meta learning.

\subsection{Model- and Task-Agnostic Performance Improvements over FedAvg}

In FL settings, the performance drop compared with optimizations in the data center is mainly caused by the non-IID data distributions.

FedShare~\cite{zhao2018federated} demonstrates the weight divergence on clients through both experiments and mathematical analysis. Then they propose to alleviate the problem by creating a small set of data samples that is shared among all the clients and servers.
Nevertheless, sharing data brings extra transmission costs and privacy concerns.
FedProx~\cite{li2018federated} uses a proximal term to improve the performance of FedAvg in heterogeneous networks.

Jeong et al. \cite{jeong2018communication} propose federated distillation~(FD), which creates prototypical vectors for each label and takes them as the substitutes for the missing data samples.
They further propose federated augmentation~(FAug) by training a generative model on the server using a small set of data.
Then each client can download the generative model and locally generate the missing data samples, converting local data distributions to IID.
However, training generative models for complex datasets is still an open problem, so the application of FD and FAug is very limited in practice. 

\subsection{Model Agnostic Meta Learning}

Model-agnostic meta learning~(MAML)~\cite{finn2017model} and its following work~\cite{li2017meta,antoniou2018train} propose fast adaption training for few-shot learning tasks.
Their basic idea is to find an initial parameter $\omega_0$, after a small number of updating steps on the support set to obtain $\omega_n$, the network parameter that performs well on the target set.
The training process using the keep-trace gradient descent on the support set is called the inner-loop update in MAML, while the optimization of the meta-objective on the target set is called the outer-loop update.
This strategy for updating $\omega_0$ is similar to the proposed UGA.
The difference lies in that we evaluate the gradients against $\omega_0$ on the same dataset as the keep-trace gradient descent, instead of on target set and support set separately.
Besides, UGA is performed on multiple clients, and we additionally introduce a controllable meta training process after model aggregation on the server.
According to the criteria in MAML, FedMeta w/ UGA is actually considered as a three-stage optimization in a sense with the first two on clients and the last on the server.

Recently, some works \cite{jiang2019improving,fallah2020personalized} have also explored the combination of meta learning and FL.
However, they mainly focus on the personalization of federated models, which distinguishes from our work.
An interesting note is that the Per-FedAvg proposed in \cite{fallah2020personalized} can be seen as a special case of our UGA with just one step keep-trace gradient descent, followed by another step of gradient evaluation.

\section{Conclusion}

In this paper, we take a closer look at the FedAvg algorithm through theoretical analysis and point out two major deficiencies.
To tackle them, we propose our model- and task-agnostic improvements, i.e., the unbiased gradient aggregation with the keep-trace gradient descent and gradient evaluation strategy, and the meta updating procedure with a controllable meta training set, which can be used separately, as well as together to achieve further improvements.
And they can be integrated into existing FL system easily.
Experimental results demonstrate that the proposed methods are faster in convergence and achieve higher accuracy than popular FL algorithms with different network architectures in various FL settings.

